\documentclass[authoryear,preprint,12pt]{elsarticle}
\usepackage[a4paper, left=1in, right=1in, top=1in, bottom=1in]{geometry}
\usepackage{amssymb}
\usepackage{amsmath}
\usepackage{enumitem}
 \usepackage{amsthm}
\usepackage{xcolor}
\usepackage{multirow}
\usepackage{multicol}
\usepackage{arydshln}
\usepackage[hyphens]{url}
\usepackage[ruled,vlined]{algorithm2e}
\usepackage[colorlinks=true, breaklinks=true]{hyperref}
\usepackage{lineno}

\journal{ISPRS Journal}

\begin{document}

\begin{frontmatter}



\title{L2M-Reg: Building-level Uncertainty-aware Registration of Outdoor LiDAR Point Clouds and Semantic 3D City Models}

%% \author[label1,label2]{Ziyang Xu}

\author[label1]{Ziyang Xu*}
\ead{ziyang.xu@tum.de}

\author[label2]{Benedikt Schwab}

\author[label1]{Yihui Yang*}
\ead{yihui.yang@tum.de}

\author[label2]{Thomas H. Kolbe}

\author[label1]{Christoph Holst}

\cortext[cor*]{Corresponding author.}

\affiliation[label1]{organization={Chair of Engineering Geodesy, TUM School of Engineering and Design, Technical University of Munich},%Department and Organization 
            city={Munich},
            postcode={80333}, 
            country={Germany}}

\affiliation[label2]{organization={Chair of Geoinformatics, TUM School of Engineering and Design, Technical University of Munich},
            city={Munich},
            postcode={80333}, 
            country={Germany}}

%% Abstract
\begin{abstract}
Accurate registration between LiDAR (Light Detection and Ranging) point clouds and semantic 3D city models is a fundamental topic in urban digital twinning and a prerequisite for downstream tasks, such as digital construction, change detection, and model refinement. However, achieving accurate LiDAR-to-Model registration at the individual building level remains challenging, particularly due to the generalization uncertainty in semantic 3D city models at the Level of Detail 2 (LoD2). This paper addresses this gap by proposing L2M-Reg, a plane-based fine registration method that explicitly accounts for model uncertainty. L2M-Reg consists of three key steps: establishing reliable plane correspondence, building a pseudo-plane-constrained Gauss-Helmert model, and adaptively estimating vertical translation. Overall, extensive experiments on five real-world datasets demonstrate that L2M-Reg is both more accurate and computationally efficient than current leading ICP-based and plane-based methods. Therefore, L2M-Reg provides a novel building-level solution regarding LiDAR-to-Model registration when model uncertainty is present. The datasets and code for L2M-Reg can be found: \url{https://github.com/Ziyang-Geodesy/L2M-Reg}.
\end{abstract}


\begin{highlights}
\item A plane-based LiDAR-to-Model fine registration method, L2M-Reg, tailored for individual buildings is proposed.

\item Explicitly considering the uncertainty in LoD2 models, L2M-Reg achieves superior performance on five real-world datasets.

\item A 2D-3D decoupled transformation estimation is introduced to mitigate the adverse impact of low-quality ground model data.

\item A lightweight plane correspondence strategy is developed that leverages the embedded semantic information in LoD2 models.
\end{highlights}

%% Keywords

\end{frontmatter}


%\begin{linenumbers}
\section{Introduction}
\label{sec1}
%% Labels are used to cross-reference an item using \ref command.
LiDAR (Light Detection and Ranging) point clouds and semantic 3D city models are two widely used digital representations that play important roles in urban digital twinning  \citep{ketzler2020,jeddoub2023}. However, as heterogeneous data types, they differ in many aspects such as data format, acquisition methods, geometric features, and accuracy. Specifically, semantic 3D city models primarily contain generalized structural information and serve as simplified and abstract representations of physical buildings, whereas LiDAR point clouds accurately capture detailed geometric features, providing high-precision representations  \citep{kada2009,Xu2021}. Accurate and efficient registration between them constitutes a fundamental research topic, as well as an essential prerequisite for numerous downstream applications such as digital construction, change detection, model updating and enrichment, model reconstruction, and texturing  \citep{Wysocki2021,Liu2024,Shao2024,Zhu2024,kulmer2025}.

Internationally, CityGML (City Geography Markup Language), as a standard developed by the Open Geospatial Consortium (OGC), is widely adopted for representing and managing semantic 3D city models \citep{groger2012,kolbeOGCCityGeography2021}. CityGML supports the modeling of urban objects by incorporating their 3D geometry, appearance, topology, and semantic information across four distinct Levels of Details (LoD) \citep{kolbeOGCCityGeography2021}. Benefiting from the widespread adoption of CityGML, accurately georeferenced LoD2 models are publicly available in many countries and regions \citep{wysocki2024}. These models are typically maintained by governmental agencies or professional institutions to ensure consistent quality and reliability. Consequently, in most LiDAR-to-Model registration tasks, existing georeferenced LoD2 models are utilized as reference data, to which up-to-date LiDAR point clouds are aligned.

\begin{figure}[hbtp]
    \centering
    \includegraphics[width=\linewidth]{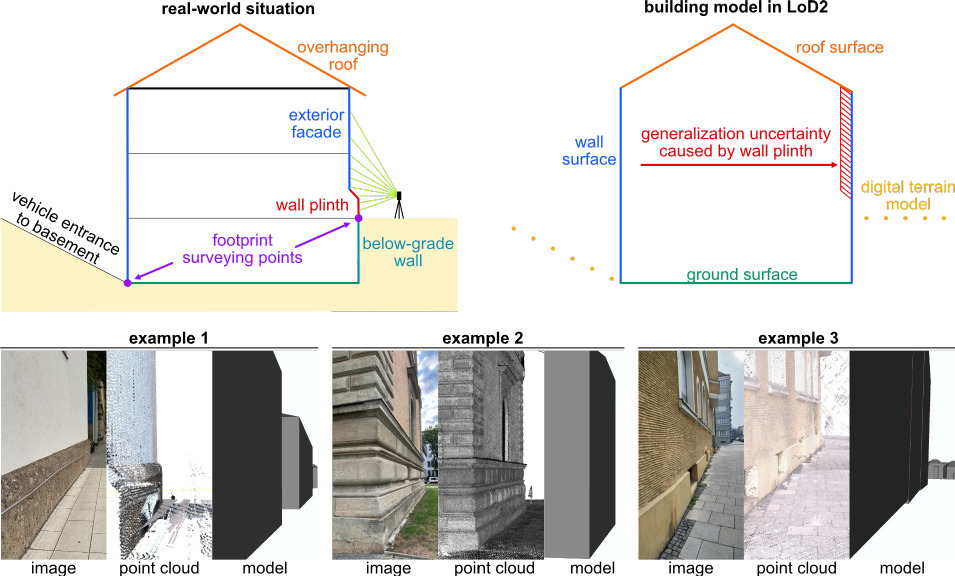}
    \caption{Sources of uncertainty in LoD2 building models. Model uncertainty primarily arises from the model generation process. In reality, building footprint points correspond to the wall plinth rather than the upper facade, resulting in a horizontal offset between the two (as indicated by the red dashed area). Since LoD2 models are generated directly from these footprint points, the modeled wall surfaces align geometrically with the plinth, but not with the actual facade above.} \label{model uncertainty}
\end{figure}

When discussing LoD2 building models used for accurate registration, a particularly critical issue is their inherent model uncertainty. In this context, model uncertainty does not refer to the traditional statistical or probabilistic interpretation, such as variance or covariance describing the positional accuracy of LoD2 model coordinates. The \emph{model uncertainty} is defined as the inconsistency between the final model, produced through a predefined generation process, and the actual building object, specifically in terms of geometric and structural representation. The encountered uncertainty stems from the current model generation process, which typically relies on 2D building footprints from the cadastral registry. As illustrated in Figure \ref{model uncertainty}, the building footprint is defined by the outermost structural elements. Therefore, it is important to clarify that this study targets structural or generative inconsistencies between LoD2 models and real-world facades rather than statistical error propagation modeled via covariance matrices.

As seen in Figure \ref{model uncertainty}, buildings typically consist of various structural elements from the foundation to the roof, such as wall plinth (in red), exterior facade (in blue), and overhanging roof (in orange). Cadastral surveying regulations define which of these elements should be included in the model. Typically, a horizontal offset exists between a building’s plinth and its upper facade. Depending on the architectural style, this offset can range from several centimeters to decimeters. Since LoD2 building models are generally reconstructed by extruding building footprints to prototypical roof shapes from Airborne Laser Scanning (ALS) data \citep{roschlaubINSPIREKONFORM3DBUILDING2016}, the modeled wall surfaces actually only align with the building's plinths, not exterior facades. Consequently, the building plinths that closely correspond to the footprint should be more suited for establishing correspondences with modeled wall surfaces.

Previous studies on LiDAR-to-Model registration have commonly assumed that the LoD2 models served as reference data are error-free, thus neglecting this inherent uncertainty. Although this assumption is basically accepted, it will become problematic in high-precision application scenarios. For example, when the captured LiDAR point clouds need to be accurately georeferenced based on existing LoD2 models, this horizontal offset between the plinth and exterior facade cannot be disregarded. Recent research has acknowledged similar issues about model uncertainty, focusing primarily on visual and conceptual representations from architectural or modeling perspectives \citep{potter2012,landes2019,zou2022}. However, most LiDAR-to-Model registration studies have overlooked this uncertainty issue, treating it as a conventional point cloud registration task. As a result, a disconnection arises between building modeling and LiDAR-to-Model registration research.

Since urban dwellings form the backbone of any city, most existing semantic 3D city models are composed of buildings \citep{biljecki2015}. Compared to city-level LiDAR-to-Model registration, an accurate and efficient building-level solution offers the potential to refine city models based on specific needs, thereby avoiding extensive data acquisition and substantially reducing costs. However, this transition introduces several challenges:

\begin{enumerate}[label=(\arabic*)]
\item \textbf{Increased accuracy requirements}. Inherent model uncertainty that is negligible at the city scale needs to be explicitly addressed at the building level for high-precision applications.
\item \textbf{Limited availability of geometric features}. Individual buildings or partial building segments typically contain fewer geometric primitives, such as facade planes, with limited diversity in orientation and quantity. This constraint makes feature extraction and correspondence establishment more difficult.
\item \textbf{Demand for balanced accuracy and efficiency}. While high accuracy is essential, efficient processing at the building level is equally important to enable scalability for city-scale applications.
\end{enumerate}

These challenges underscore a clear research gap in achieving accurate building-level LiDAR-to-Model registration under model uncertainty. To address this, a tailored solution named L2M-Reg is proposed. To the best of our knowledge, this is the first comprehensive study to explicitly account for the inherent uncertainty of reference LoD2 models in building-level registration. The main contributions are summarized as follows:

\begin{enumerate}[label=(\arabic*)]
\item A plane-based LiDAR-to-Model fine registration method, L2M-Reg, tailored for individual buildings is proposed. It integrates reliable plane correspondence establishment, pseudo-plane-constraint Gauss-Helmert Model (GHM), and adaptable vertical translation estimation. By explicitly considering the uncertainty in LoD2 models, L2M-Reg achieves superior performance on all five real-world datasets.

\item A 2D-3D decoupled transformation parameter estimation strategy is introduced to mitigate the adverse impact of low-quality ground model data on the overall accuracy of 6 Degree-of-Freedom (DoF) parameter estimation. By decoupling vertical and horizontal components, this strategy effectively prevents high ground model uncertainty from degrading horizontal registration accuracy.

\item A lightweight plane correspondence strategy is developed that leverages the embedded semantic information in LoD2 models. It eliminates the conventional need for converting models into point clouds and performing feature-based matching, thereby significantly improving robustness and computational efficiency. Furthermore, it is built upon the internationally recognized CityGML standard, ensuring high interoperability and ease of adoption across different countries and regions.
\end{enumerate}

This paper starts with a background introduction on LiDAR-to-Model registration and the inherent uncertainty of LoD2 models, followed by a detailed review of related works in Section~\ref{sec2}. Section~\ref{sec3} presents the detailed methodology of L2M-Reg. Section~\ref{sec4} describes the experimental results. Section~\ref{sec5} discusses the advantages and limitations, followed by conclusions and outlook in Section~\ref{conclusion}.

\section{Related Works}
\label{sec2}

This paper focuses on accurate LiDAR-to-Model registration while explicitly accounting for the inherent uncertainty in LoD2 models. Thus, the related work is divided into LiDAR-to-Model registration (Section \ref{subsec2.1}) and the inherent uncertainty of models (Section \ref{subsec2.2}).

\subsection{LiDAR-to-Model Registration}
\label{subsec2.1}
Accurate registration between LiDAR point clouds and LoD2 models is an essential and fundamental task. In recent years, researchers worldwide have extensively investigated this topic, which can be summarized into two categories: registration directly using LoD2 models in Section \ref{subsubsec2.1.1} and registration using converted point clouds from LoD2 models in Section \ref{subsubsec2.1.2} \citep{bueno2018,sheik2022a}.

\subsubsection{Registration Directly Using LoD2 Models}
\label{subsubsec2.1.1}

The registration between LiDAR point clouds and their corresponding LoD2 models essentially involves feature extraction and correspondence matching across heterogeneous data. A straightforward strategy to address this issue is to directly extract geometric features, such as points, lines, and planes, from the models and subsequently match these features with corresponding ones derived from the point clouds  \citep{bosche2012,bueno2018,goebbelsITERATIVECLOSESTPOINT2019,lucksImprovingTrajectoryEstimation2021,sheik2022a,sheik2022b,monasse2023}. The transformation parameters can then be computed based on these matched features. In particular, most man-made structures include numerous planar surfaces, which naturally facilitates the extraction and utilization of planar features \citep{sheik2022a,zhao2022,qiao2023,yang2023towards,zhang2024l}.

Another strategy is to use the footprint points or polygons of the entire model as features to calculate the rotation and translation parameters  \citep{diakite2020}. However, methods relying on footprint points or polygon-based strategies encounter difficulties when dealing with symmetrical buildings, as these methods cannot accurately estimate the rotation parameters of entirely symmetrical structures and may introduce ambiguities.

It should be noted that the registration strategy directly utilizing LoD2 models is feasible, with its primary advantage being that it does not require converting the models into intermediate point clouds, thus making the process concise and efficient. However, a major challenge of this strategy is its dependence on the quantity and quality of features extracted from the models. If the models themselves lack sufficient features, this deficiency can easily result in decreased registration accuracy \citep{bosche2012, sheik2022b}. To address this issue, an increasing number of researchers are adopting methods that generate 3D point clouds from LoD2 models and convert the original LiDAR-to-Model registration into a typical point cloud registration task.

\subsubsection{Registration Using Converted Point Clouds from LoD2 Models}
\label{subsubsec2.1.2}

As mentioned in Section~\ref{subsubsec2.1.1}, a LiDAR-to-Model registration task can be converted into classic point cloud registration. Over the past few decades, extensive research regarding this topic has been conducted across multiple fields, including geodesy, GIS, computer science, robotics, AEC (Architecture, Engineering, Construction), etc., with each domain emphasizing distinct aspects like scenarios and accuracy. Given that this paper primarily addresses the fine LiDAR-to-Model registration in the context of individual buildings, the following literature review will specifically focus on two relevant categories of registration: ICP-based methods and geometric feature-based methods.

ICP-based methods are extensively employed to register point clouds. Traditional ICP algorithms iteratively align corresponding points but often exhibit slow convergence and sensitivity to initial values and outliers \citep{besl1992}. Thus, many enhanced variants have been developed to address these limitations. For instance, Point-to-Line ICP  \citep{censi2008} and Point-to-Plane ICP \citep{rusinkiewicz2001,low2004} utilize local linear and planar features, respectively, thereby accelerating convergence and enhancing registration accuracy. Generalized ICP (GICP) further refines registration precision by statistically optimizing correspondences through probabilistic modeling of local surface geometry \citep{segal2009}. Trimmed ICP (TriICP) effectively reduces the influence of outliers by selectively excluding a certain percentage of correspondences, thereby enhancing robustness in noisy scenarios \citep{chetverikov2002}. More recently, a robust method based on generated planar patches and adaptive distance thresholds was proposed, which has significantly reduced the influence of surface changes on registration accuracy \citep{yang2023, yang2025}. KISS-ICP is also widely used for point cloud registration because of its versatility and computational efficiency, particularly in robotics and SLAM applications \citep{vizzo2023}. Collectively, these ICP variants have progressively addressed the limitations in standard ICP, significantly improving the accuracy, efficiency, and practical applicability of point cloud registration tasks.

Compared with the aforementioned ICP-based methods, geometric feature-based methods utilize points, lines, curves, and planes extracted from point clouds to establish correspondences for transformation estimation. These methods are generally less sensitive to variations of initial alignments compared to ICP-based methods. In the building scenarios discussed in this paper, planar features offer notable advantages over other geometric primitives, such as point or line features, and have thus been commonly utilized in point cloud registration tasks \citep{bosche2012,Wujanz2018,chen2019,sheik2022a,xu2025pl4u}. This preference arises primarily for two reasons. First, man-made structures typically contain abundant planar elements, naturally facilitating the extraction and use of planar features. Second, planar features demonstrate higher resilience against surface noise and outliers. 

More specifically, Scantra \citep{Wujanz2018}, developed by technet GmbH\footnote{\url{https://www.technet-gmbh.com/en/products/scantra/}}, projects the point cloud onto a 2D space and leverages extracted planar information, such as area, bounding box, boundary length, and average intensity, to establish plane correspondences \citep{Dold2006}. PLADE \citep{chen2019} utilizes plane pairs and their spatial intersection lines as structural foundations to construct descriptors, ultimately achieving high-accuracy registration. Both of them are two leading plane-based methods which have been widely used in point cloud registration \citep{Holst2019,Kaiser2022,ma2023}.

According to the summary above, the majority of published studies still regard ICP-based methods and PLADE as appropriate state-of-the-art baselines for comparison in the context of LiDAR-to-Model registration.

\subsection{Inherent Uncertainty of Models}
\label{subsec2.2}

Compared to the rapid growth in the creation and application of models, research on their uncertainty has lagged behind and not received equivalent attention \citep{zou2022}. Initially, researchers from architecture and archaeology took the lead in visualizing uncertainty associated with models of historical buildings \citep{zuk2005,kensek2007}. Subsequently, solutions to quantifying uncertainty in models have continued to evolve, with increasingly comprehensive methods proposed, along with the introduction of the concept of Level of Uncertainty (LoU), which systematically characterizes uncertainty variations among different components of building models \citep{potter2012,landes2019}.

With growing demand for high-precision applications relying on semantic 3D city models, such as navigation \citep{kulmer2025}, trajectory estimation \citep{lucksImprovingTrajectoryEstimation2021}, and change detection \citep{meyer2022change}, an increasing number of researchers have also begun to investigate the uncertainty inherent in these models \citep{foschi2024}. For instance, probabilistic methods have been developed for quantifying uncertainty, primarily focusing on semantic labeling in 2D maps or building models \citep{paz2020,feng2021,di2022}. Additionally, other researchers have specifically examined the uncertainty associated with facades and windows in semantic 3D city models, quantifying aspects such as their positions and orientations \citep{zou2022}.

Collectively, uncertainty in semantic 3D city models has become increasingly important in high-precision applications, as uncertainty inherent in the model inevitably propagates into downstream use cases. Neglecting these uncertainties or treating the model as an error-free reference is likely to introduce systematic inaccuracies and errors in subsequent processes \citep{zou2022}. Specifically, within the context of accurate building-level LiDAR-to-Model registration discussed in this study, addressing uncertainty inherent in the model serving as reference data will be a central focus.

\section{Methodology} 
\label{sec3}

\begin{figure}[hbtp]
  \centering
  \includegraphics[width=1.0\linewidth]{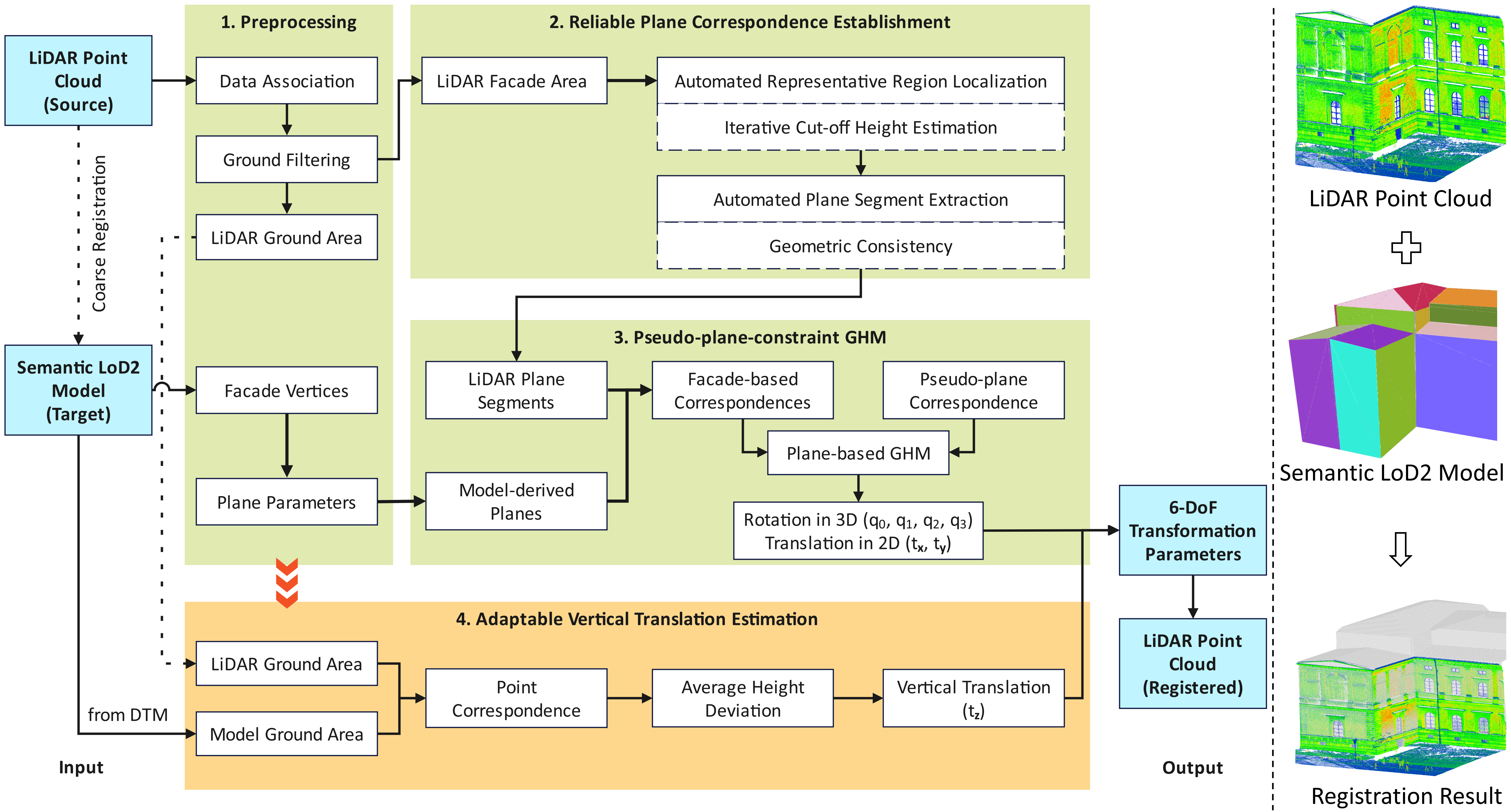}
  \caption{Flowchart of the proposed L2M-Reg. Each plane of the input semantic LoD2 model is colored for better visualization, and the point cloud is colored by intensity.}
  \label{workflow}
\end{figure}

Figure \ref{workflow} gives an overview of the proposed L2M-Reg. The required inputs are the semantic LoD2 models of the individual building and its corresponding coarsely registered LiDAR point clouds. The LiDAR point clouds could be acquired from MLS systems equipped with GNSS or other well-established coarse registration solutions. The implementation of coarse registration falls outside the scope of this study, and more details can be found from the literature \citep{xu2017automated, bueno2018, xu2019pairwise}. After preprocessing introduced in Section \ref{subsec3.1}, L2M-Reg consists of three key steps: reliable plane correspondence establishment in Section \ref{subsec3.2}, pseudo-plane-constrained Gauss-Helmert model in Section \ref{subsec3.3}, and adaptable vertical translation estimation in Section \ref{subsec3.4}. The core of the L2M-Reg lies in addressing the spatial variation in model uncertainty among different building structural components. L2M-Reg automatically localizes and extracts representative planar regions (i.e., wall plinth) to establish correspondences and obtain 6 DoF transformation parameters. More detailed steps are given in the following subsections.

\subsection{Data Preprocessing}
\label{subsec3.1}
Data preprocessing is performed on the LiDAR point clouds and corresponding LoD2 models. For outdoor LiDAR-to-Model registration, the required overlapping region between two datasets is generally limited to ground areas and exterior building facades. This is because adopted Terrestrial Laser Scanning (TLS) and Mobile Laser Scanning (MLS) systems cannot capture the roof regions.

Publicly available Digital Terrain Model (DTM) datasets can provide additional ground information, as these datasets are georeferenced and maintained by governmental or professional institutions, ensuring their high authority and accessibility. Consequently, the model information used in this study is primarily from two components: building facades and ground areas adjacent to these facades.

The LoD2 models utilized as reference data comprise multiple wall surfaces with planar geometry representation. Each wall acts as a reference plane for subsequent registration and possesses a unique object ID along with the corresponding coordinates of several vertices, as specified by the CityGML standard. These vertices are employed to calculate the associated plane parameters of each model.

\begin{figure}[hbtp]
    \centering
    \includegraphics[width=\linewidth]{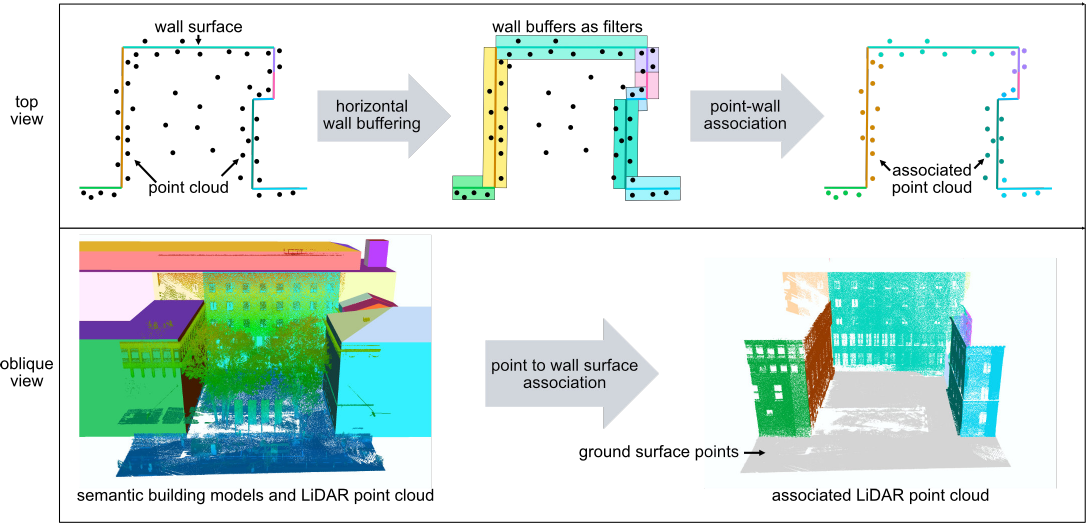}
    \caption{Data association illustration for LiDAR point clouds and wall surfaces. To establish correspondence, only points located within each wall’s buffer zone are retained. The resulting associations are colorized to facilitate clearer interpretation.}\label{data association}
\end{figure}

Since the Z-axis of LiDAR point clouds is adjusted to be vertically upward by the sensors, all wall surfaces are assumed to be extruded along their normal vectors in horizontal directions, as shown in Figure \ref{data association}. The resulting buffer zones with a thickness of 1 m are used to associate individual points with the wall surfaces within whose buffers they reside. The choice of 1 m is mainly based on two considerations: first, many existing LiDAR-to-Model coarse registration methods already achieve decimeter-to-centimeter-level accuracy \citep{diakite2020, sheik2022a, sheik2022b}; second, as building plinths are generally within the centimeter-to-decimeter scale, a 1 m buffer reliably covers them without adding processing overhead. Similarly, the DTM is converted into a triangulated irregular network and extruded to filter out redundant LiDAR data. This process also removes elements such as vegetation and pedestrians located near buildings.

Simultaneously, during the buffer zone-based filtering, each individual point is associated with the corresponding wall surface by assigning them unique wall identifiers, as illustrated in Figure \ref{data association}. Although the finally usable plane segments are not explicitly extracted at this stage, they must reside within the associated LiDAR point clouds. Accordingly, the correspondence relationships are firmly locked at this step. 

The built-in semantics in LoD2 models play a critical role in this step, particularly in the accurate identification of wall surfaces. These semantics allow wall surfaces to be reliably selected and associated with the corresponding LiDAR point clouds through buffer zone-based filtering. In this way, semantic information is transferred from the model to the point clouds, enabling identification of both the wall surfaces in the model and their corresponding point cloud regions. This process provides the foundation for the subsequent extraction of wall plinths and enables more reliable identification of wall-related regions without introducing additional data labeling costs.

In contrast to other plane-based methods that first extract plane segments and then search for correspondences \citep{Wujanz2018,chen2019,sheik2022a}, this strategy maximizes the utilization of existing semantic information from the models. By reliably fixing correspondence relationships before extracting specific plane segments, this strategy ensures a more effective use of prior information while achieving greater simplicity and efficiency as shown in Section \ref{subsubsection4.4.3}.

The main outputs of data preprocessing consist of the model-derived plane $M_i$ of the facade area and the corresponding neighboring point cloud $N_i$ (equal to the associated LiDAR point clouds shown in Figure \ref{data association}). The parameters of each $M_i$ are explicitly known, and stable correspondence relationships between $N_i$ and $M_i$ are fixed through data association. In the subsequent step described in Section \ref{subsec3.2}, the finally usable plane correspondences will be established.

\subsection{Reliable Plane Correspondence Establishment}
\label{subsec3.2}

After data preprocessing, accurately extracting a suitable LiDAR plane segment $L_i$ from neighboring point cloud $N_i$ to construct reliable correspondences remains a challenge. To address this, an automated solution is developed, comprising two main parts: representative region localization in Section \ref{subsubsec3.3.1} and plane segment extraction in Section \ref{subsubsec3.3.2}. In general, the input of this step consists of the point clouds covering the entire facade region, including parts of the overhanging roof, the exterior facade, and, most importantly, the wall plinth, as shown in Figure \ref{model uncertainty}. The output is the extracted plane segment $L_i$ (wall plinth area). Since the wall plinth is the primary source of the model uncertainty addressed in this study.

\subsubsection{Automated Representative Region Localization}
\label{subsubsec3.3.1}

In this part, the input consists of the neighboring point cloud $N_i$ surrounding each model-derived plane $M_i$, encompassing the full exterior facade area of the building. As noted before, previous plane-based LiDAR-to-Model registration methods often overlook the model uncertainty and directly use the most representative plane from the full neighboring space to establish correspondences. In this case, they implicitly assume that the planes derived from the entire $N_i$ are inherently representative. However, this assumption becomes invalid when considering model uncertainty.

Specifically, in LoD2 models, wall surfaces are typically generated by vertically extruding the building footprint, implying that the model-derived plane $M_i$ should only correspond to the building's plinth, as shown in Figure \ref{model uncertainty}. This plinth region is actually a subspace of each neighboring point cloud $N_i$ and contains the representative region used for correspondence construction. Therefore, the goal in this step is to automatically localize this representative region. Herein, an iterative cut-off height estimation algorithm is developed to eliminate unsuitable regions and adaptively localize the plinth areas for subsequent LiDAR plane segment $L_i$ extraction. The core of this algorithm lies in the continuous refinement of the neighboring space. As shown in Figure \ref{iterative cut-off}, the input neighboring point cloud is $N_i$, where $i$ denotes the shared ID of the associated model-derived plane $M_i$. The algorithm is conducted based on the following steps:

\begin{figure}[hbtp]
\centering
\includegraphics[width=1.0\linewidth]{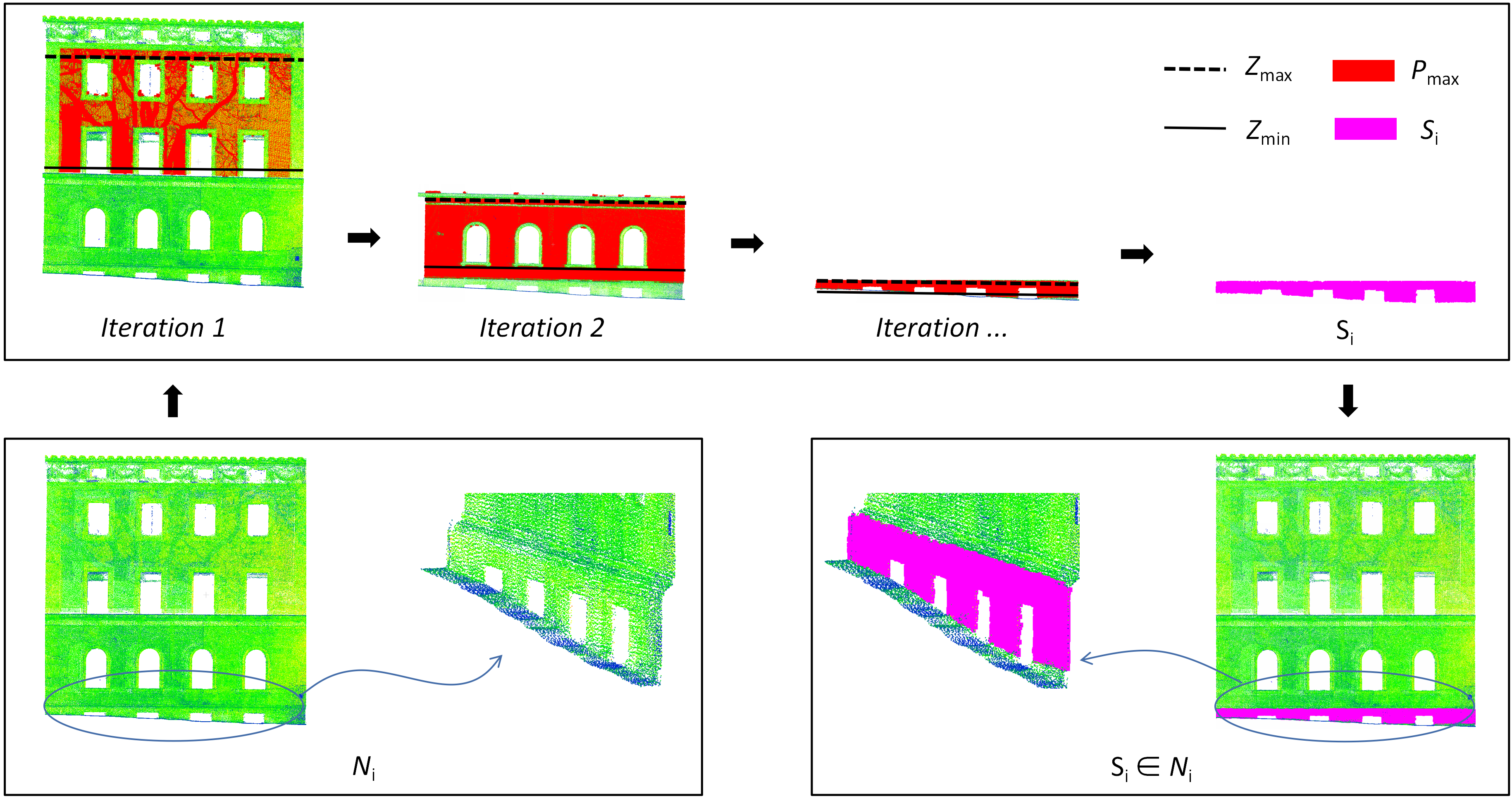}
\caption{Automated representative region localization. $N_i$ represents the input neighboring point cloud (colored by intensity), and $S_i$ represents the output representative subspace (in purple) corresponding to the building's plinth.}\label{iterative cut-off}
\end{figure}

\begin{enumerate}[label=(\arabic*)]
    \item Detecting the largest plane from $N_i$ based on Random Sample Consensus (RANSAC) as shown in the red part in Figure \ref{iterative cut-off}, denoted as $P_{\text{max}}$. The distance threshold $T_{{dis}}$ used in the RANSAC is determined based on the registration residual $e$ of the entire input LiDAR point clouds, defined as $T_{{dis}} = e$. The residual $e$ can be derived from the performance of the multi-station point cloud registration.

    \item The $Z$ coordinates of all the points in $P_{\text{max}}$ are then sorted in ascending order, and the value at the 10th and 90th percentiles is selected as the lower and upper height bounds as shown in solid and dashed black lines in Figure \ref{iterative cut-off}, respectively, denoted as $Z_{\min}$ and $Z_{\max}$.
    
    \item The angle $\alpha$ between the normal vector of $P_{\text{max}}$ and the ground plane is then calculated. If $\alpha$ exceeds angle threshold $T_{{\alpha}}$, $P_{\text{max}}$ is considered part of the exterior facade. In this case, all points in $N_i$ whose $Z$ coordinates are larger than $Z_{\min}$ are removed, and the $P_{\text{max}}$ extraction process is also recognized as valid. The remaining point cloud is then taken as the updated $N_i$ for the next iteration, as shown in Figure \ref{iterative cut-off}.  
\end{enumerate}

The above steps are repeated until the extraction of $P_{\text{max}}$ is deemed invalid, indicating that all extractable instances of $P_{\text{max}}$ have been traversed from top to bottom along the height direction. The angular threshold $T_{{\alpha}}$ is set to a predefined value of $10^\circ$, which is empirically determined based on the structural characteristics of typical buildings. The last validly extracted $P_{\text{max}}$ is then selected, and its corresponding $Z_{\min}$ and $Z_{\max}$ values are computed. These values are used to define the final cut-off height range $R$ based on

\begin{equation}
R \in [Z_{\min}, Z_{\max}].
\end{equation}

Only the points in $N_i$ whose $Z$-coordinates fall within this range $R$ are retained, forming the final output representative subspace $S_i$. The subspace $S_i$ primarily includes the building's plinth, which serves as the representative region for the subsequent LiDAR plane segment $L_i$ extraction input. The main steps for the iterative cut-off height estimation are shown in Algorithm \ref{Algorithm1}.

\begin{algorithm}[htbp]
\caption{Iterative Cut-off Height Estimation for Representative Subspace}
\label{Algorithm1}
\KwIn{Neighboring point cloud $N_i$}
\KwOut{Representative subspace $S_i$}

Set angular threshold $T_\alpha$ \;
Set distance threshold $T_{\text{dis}}$ \;
\Repeat{extraction of $P_{\text{max}}$ is invalid}{
    Detect largest plane $P_{\text{max}}$ from $N_i$ using RANSAC with threshold $T_{\text{dis}}$ \;
    Sort $Z$ coordinates of points in $P_{\text{max}}$ ascendingly \;
    Compute $Z_{\min} \leftarrow 10$th percentile, $Z_{\max} \leftarrow 90$th percentile \;
    Compute angle $\alpha$ between normal of $P_{\text{max}}$ and ground plane \;
    \If{$\alpha > T_\alpha$}{
        Remove all points in $N_i$ where $Z > Z_{\min}$ \;
        Update $N_i$ with remaining points \;
        Mark extraction as valid \;
    }
    \Else{
        Mark extraction as invalid \;
    }
}
Let last valid $P_{\text{max}}$ define $Z_{\min}, Z_{\max}$\;
Define cut-off height range $R \leftarrow [Z_{\min}, Z_{\max}]$ \;
Retain points in $N_i$ where $Z \in R$ to form final subspace $S_i$ \;

\Return $S_i$
\end{algorithm}

\subsubsection{Automated Plane Segment Extraction}
\label{subsubsec3.3.2}

After obtaining the subspace $S_i$ from $N_i$, the next task is to extract the desired LiDAR plane segment $L_i$ from $S_i$ to establish a correspondence with $M_i$. It is worth noting that the actual point cloud of $N_i$ is often incomplete due to occlusions caused by vegetation, vehicles, pedestrians, and other objects during data acquisition. Additionally, the facades and plinths of some buildings may be relatively rough and contain various irregular microstructure elements. Under such conditions, direct plane extraction typically results in a large number of discrete planes oriented in different directions.

The desired LiDAR plane segment $L_i$ should be as representative as possible and accurately reflect the location of the building’s footprint. To accomplish this, an automated extraction algorithm is developed, with the main steps outlined as follows:

\begin{enumerate}[label=(\arabic*)]
    \item Plane extraction is performed on the $S_i$ using RANSAC, with the same distance threshold $T_{{dis}}$ defined previously. For each extracted plane, the angle between its normal vector and the ground plane is calculated. Only those planes with angles exceeding $T_{{\alpha}}$ are retained as candidate planes.

    \item The candidate planes are then clustered based on the similarity of their normal vectors, using a stricter angular threshold defined as $T_{\theta} = 0.5 T_{\alpha}$ to further group geometrically similar planes. Subsequently, the planes within each cluster are merged to form unified planar regions.

    \item The largest merged plane is selected as the initial seed plane and then gradually extended by incorporating candidate points from other merged planes that exhibit geometric consistency (GC). To formally define GC, let the current seed plane be denoted by \( \omega_0 \), expressed in the general form:
\begin{equation}
\omega_0: \mathbf{n}_0 \cdot \mathbf{x} + d_0 = 0
\label{eq:plane-equation}
\end{equation}

where \( \mathbf{n}_0 \in \mathbb{R}^3 \) is the unit normal vector of the seed plane, \( d_0 \in \mathbb{R} \) is the offset term, and \( \mathbf{x} \in \mathbb{R}^3 \) represents a point in 3D space. Given a candidate point \( \mathbf{p} = (x_p, y_p, z_p)^T \), its perpendicular distance to the seed plane is computed as:

\begin{equation}
d(\mathbf{p}, \omega_0) = \left| \mathbf{n}_0 \cdot \mathbf{p} + d_0 \right|.
\label{eq:point-to-plane}
\end{equation}

If \( \mathbf{p} \) is temporarily added and a new plane is re-estimated, let \( \mathbf{n}_{\text{new}} \) denote the updated unit normal vector. The angular deviation between the original and updated planes is then given by:
\begin{equation}
\theta = \arccos\left( \mathbf{n}_0 \cdot \mathbf{n}_{\text{new}} \right).
\label{eq:normal-angle}
\end{equation}

A candidate point \( \mathbf{p} \) is accepted by \( \omega_0 \) if both the distance and angular consistency criteria are satisfied:
\begin{equation}
GC(\mathbf{p}, \omega_0) =
\begin{cases}
\text{true}, & \text{if } d(\mathbf{p}, \omega_0) < T_{\text{dis}} \text{ and } \theta < T_{\theta} \\
\text{false}, & \text{otherwise}.
\end{cases}
\label{eq:geometric-continuity}
\end{equation}

    \item The extended plane obtained in step (3) is used as the final candidate point set. The plane is then re-fitted to this set using least-squares, and points within the $T_{{dis}}$ are retained as the desired LiDAR plane segment $L_i$, effectively eliminating residual noise.
\end{enumerate}

More details are shown in Algorithm \ref{Algorithm2}. The proposed plane extraction algorithm is better suited than the classic RANSAC to extract representative planes from irregular and rough areas in real-world scenarios. 

\begin{algorithm}[htbp]
\caption{Automated Plane Segment Extraction}
\label{Algorithm2}
\KwIn{Reliable subspace $S_i$, $T_{\text{dis}}, T_{\alpha}$}
\KwOut{Desired LiDAR plane segment $L_i$}

Initialize candidate plane set $\mathcal{P} \leftarrow \emptyset$ \;
Perform RANSAC-based plane extraction on $S_i$ \;
\ForEach{extracted plane $P$}{
    Compute angle $\alpha$ between normal of $P$ and ground plane \;
    \If{$\alpha > T_{\alpha}$}{
        Add $P$ to $\mathcal{P}$ \;
    }
}
Cluster planes in $\mathcal{P}$ based on normal vector similarity using $T_\theta \leftarrow 0.5 T_\alpha$ \;
Merge planes within each cluster into unified planar regions \;

Select the largest merged plane as seed plane $\omega_0$ with normal $\mathbf{n}_0$ and offset $d_0$ \;

Initialize point set $Q \leftarrow$ points in $\omega_0$ \;
\ForEach{candidate point $\mathbf{p}$ from other merged planes}{
    Compute point-to-plane distance $d(\mathbf{p}, \omega_0)$ using Eq.~\eqref{eq:point-to-plane} \;
    Temporarily add $\mathbf{p}$ to $Q$ and re-estimate plane $\omega_{\text{new}}$ with normal $\mathbf{n}_{\text{new}}$ \;
    Compute angular deviation $\theta$ using Eq.~\eqref{eq:normal-angle} \;
    \If{$d(\mathbf{p}, \omega_0) < T_{\text{dis}}$ \textbf{and} $\theta < T_{\theta}$}{
        Accept $\mathbf{p}$ into $Q$ based on geometric consistency (Eq.~\eqref{eq:geometric-continuity}) \;
        Update $\omega_0$ with new estimate \;
    }
}

Refit plane to $Q$ using least-squares \;
Retain all points in $Q$ within distance $T_{\text{dis}}$ to obtain final LiDAR plane segment $L_i$ \;

\Return $L_i$
\end{algorithm}

\begin{figure}[hbtp]
    \centering
    \includegraphics[width=0.65\linewidth]{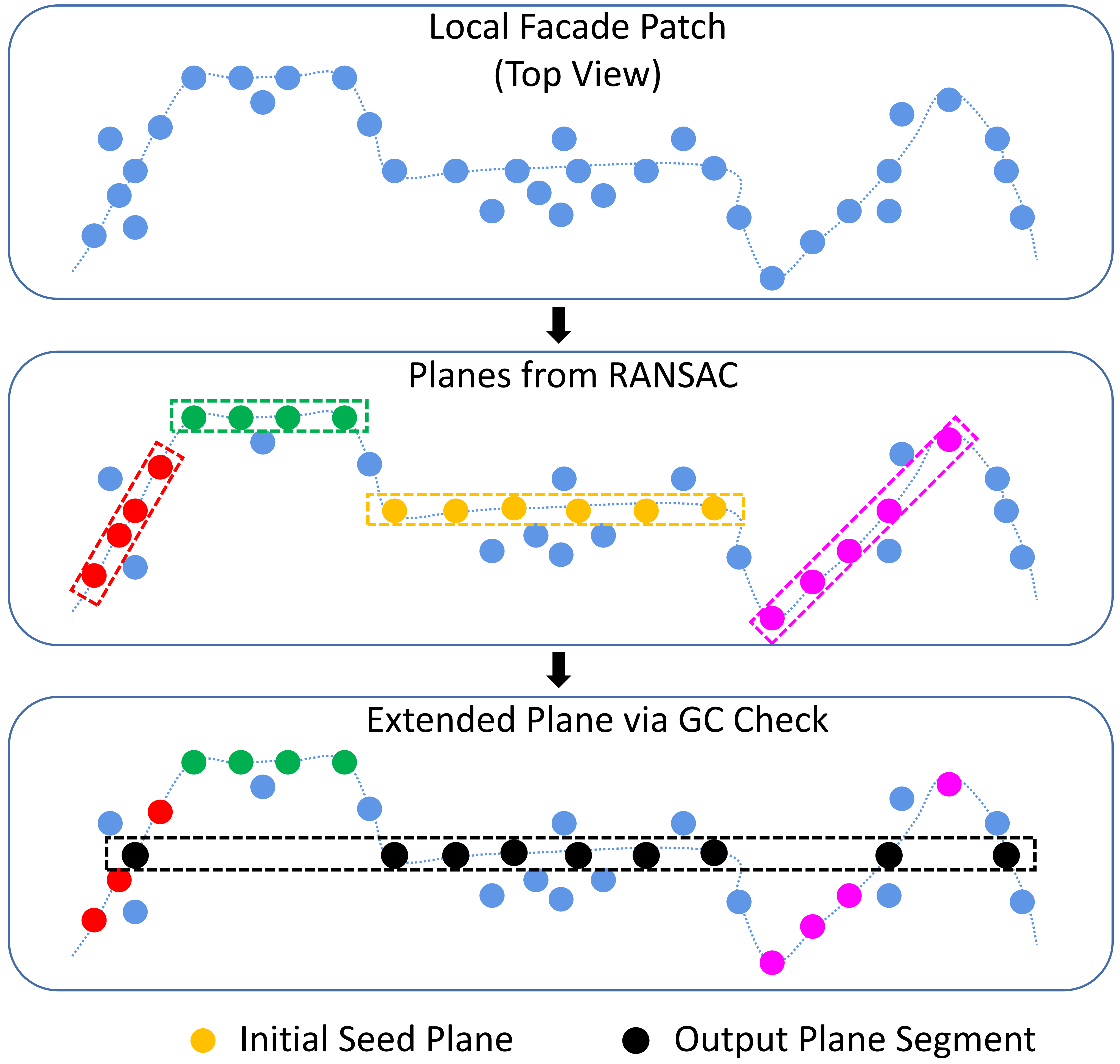}
    \caption{Desired LiDAR plane segment extraction based on geometric consistency (GC) from local facade patch. Blue points indicate a rough facade patch at a local scale. Distinct planes extracted and merged using RANSAC are shown in different colors, while black highlights the plane segments extended after the GC check, representing more complete and representative structures.}\label{GCcheck}
\end{figure}

Typically, RANSAC identifies a set of inliers that satisfy a predefined threshold. It outputs the plane corresponding to the largest inlier set while discarding all outliers by default. However, in practice, the building surface may contain many small-scale irregularities or exhibit non-planar components. In addition, due to registration residual introduced by multi-station scanning, the input point clouds used for plane extraction may suffer from local layer separation. In such cases, simply using RANSAC for plane extraction yields suboptimal results, producing multiple discrete planes. Some of the points rejected by RANSAC as outliers may, in fact, still be suited and usable. By leveraging GC check as shown in Figure \ref{GCcheck}, the proposed algorithm supplements the inlier set with structurally coherent points from other sets and outliers. More specifically, the proposed plane extraction algorithm extends RANSAC with similarity-based clustering and merging steps. The GC check is then applied to further improve the reliability of the extracted planes. As a result, the method remains applicable when exterior facades contain partially non-planar components.

After the above process is completed, the output is the final LiDAR plane segment $L_i$, whose corresponding model-derived plane $M_i$ has already been fixed during preprocessing. However, it is important to note that all current plane correspondences only cover the building exterior facade or plinth area. To achieve the 6 DoF parameter estimation, additional correspondence from the ground area is still required. This issue will be addressed in the following Section \ref{subsec3.3}.

\subsection{Pseudo-plane-constraint Gauss-Helmert Model}
\label{subsec3.3}

Through previous steps, all facade-based correspondences are available. However, ground-plane correspondences are still missing. Unlike facade areas that typically offer sufficient geometric information through well-defined vertical structures, ground areas are often underrepresented in LoD2 models. The commonly used strategy is to extract ground planes directly from publicly available DTM data, which are then combined with facade-based correspondences to jointly estimate the full transformation \citep{kumar2019,schuegraf2024}. This strategy assumes that DTM data accurately represent the true ground surface and can reliably constrain vertical translation.

\begin{figure}[hbtp]
\centering
\includegraphics[width=1.0\linewidth]{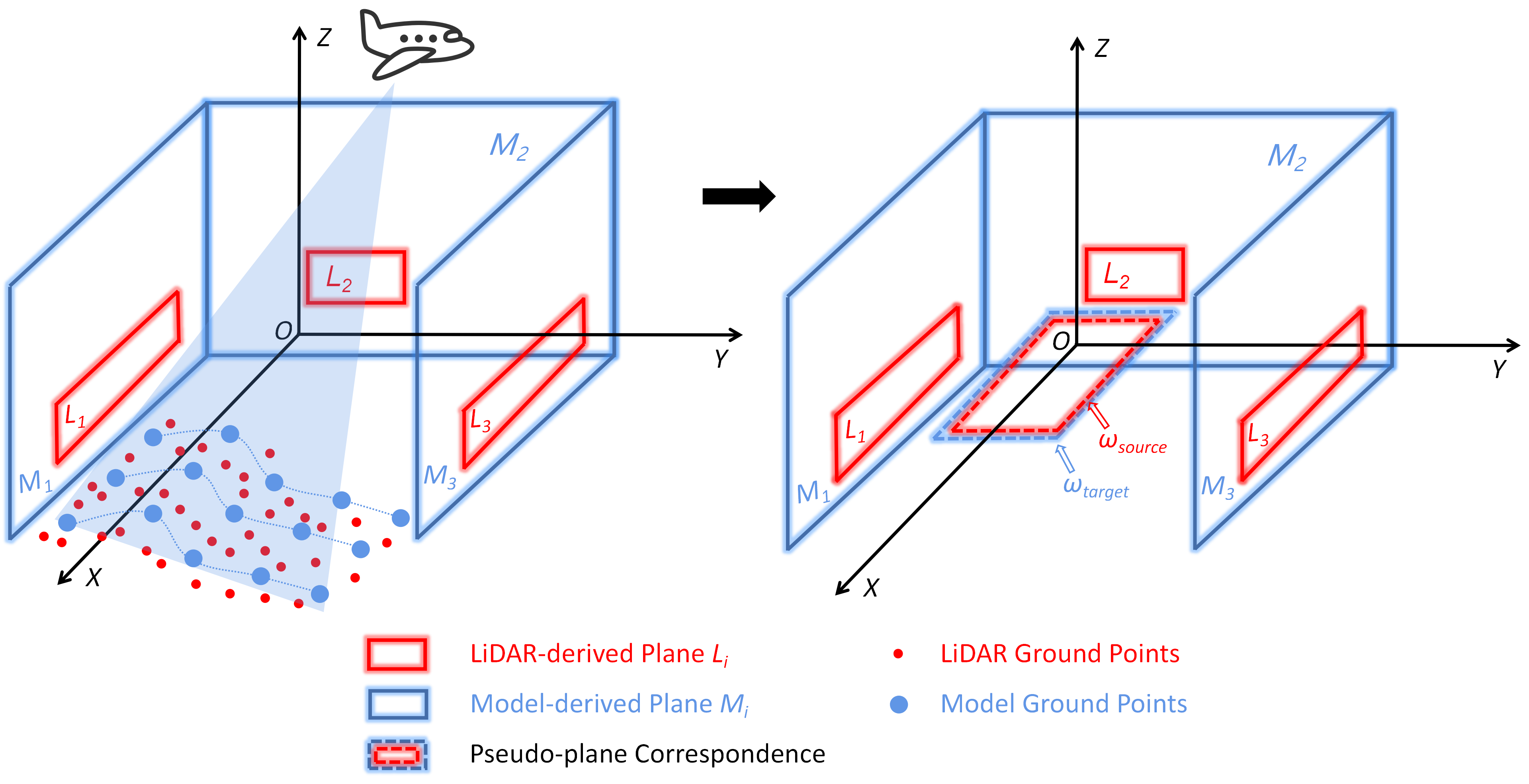}
\caption{Introduced pseudo-plane correspondence. \( L_i \) and \( M_i \) denote the LiDAR-derived and model-derived planes, respectively, and serve as facade-based correspondences for registration. The dashed lines represent the introduced pseudo-plane correspondence, which serves as a substitute for the original constraints derived from the LiDAR ground points and model ground points (DTM), indicated by red and blue round dots, respectively.}\label{pseudo-plane}
\end{figure}

However, in practice, the effectiveness of DTM data is often limited. Most DTM data are derived from Airborne Laser Scanning (ALS), which provides significantly lower resolution and geometric accuracy than the TLS or MLS data used for facade-based correspondences \citep{macay2013}, as shown in Figure \ref{pseudo-plane}. As a result, the relatively low vertical accuracy of DTM-derived correspondences will introduce uncertainty into the overall transformation estimation. In particular, errors in the vertical direction may propagate and adversely affect the estimation of other components, such as horizontal translation and rotation, ultimately compromising the accuracy and stability of the full parameter estimation. This is primarily due to the coupled nature of estimation using plane-based Gauss-Helmert Model (GHM), where vertical inaccuracies can distort the orientation of fitted planes and bias the least-squares optimization process, thereby contaminating the estimation of parameters in other directions.

To address this, a decoupled 2D-3D estimation strategy is proposed to effectively prevent vertical inaccuracies from propagating into other components. Specifically, a pseudo-plane correspondence is introduced to replace the unreliable correspondence from DTM and provide an additional constraint in the GHM adjustment.

The pseudo-plane correspondence consists of two planes with identical normal vectors, simulating the ground surfaces in the target models and the source LiDAR point clouds, respectively. Both planes are defined as:
\begin{equation}
\omega_{\text{target}}: \quad \mathbf{n} \cdot \mathbf{x} + d_t = 0
\end{equation}

\begin{equation}
\omega_{\text{source}}: \quad \mathbf{n} \cdot \mathbf{x} + d_s = 0
\end{equation}

where \( \mathbf{n} \) is a fixed unit normal vector \(\mathbf{n} = [0, 0, 1] \) pointing vertically upward. The offset terms are also set to be equal, i.e., \( d_t = d_s = 0 \), such that the pseudo-planes are fully coincident in space. As seen in Figure \ref{pseudo-plane}, this design ensures that the pseudo-plane correspondence introduces no actual height difference and serves solely as a formal constraint to stabilize the vertical component of the transformation during estimation.

In the classic plane-based GHM, each plane correspondence contributes a constraint of the form:

\begin{equation}
f = \mathbf{n}_i (\mathbf{R} \mathbf{x} + \mathbf{t}) + d_i = 0
\label{eq:ghm_plane}
\end{equation}

where \( \mathbf{n}_i \) and $d_i$ denotes the normal vector and offset term of the model-derived plane \( M_i \), \( \mathbf{x}\) is a point on the LiDAR plane segment \( L_i \), \( \mathbf{R} \in \mathbb{R}^{3 \times 3} \) and \( \mathbf{t}=[t_{x},t_{y},t_{z}]^T\) are the rotation matrix and translation vector to be estimated, respectively.

To ensure numerical stability and avoid gimbal lock, rotation is represented using a unit quaternion \( \mathbf{q} = [q_0, q_1, q_2, q_3]^T \) with \( \|\mathbf{q}\| = 1 \). The unknown parameter vector is defined as:

\begin{equation}
\mathbf{X} = [q_0, q_1, q_2, q_3, t_x, t_y, t_z]^T.
\end{equation}

The classic constraint GHM is already well-established \citep{Mikhail1976, holst2014biased, Kalenjuk2022}. Here, the original constraint is from quaternions, as seen in Eq.~\eqref{eq:c}. Three Jacobian matrices of the constraint GHM are calculated as shown in Eq.~\eqref{eq:ABC}:

\begin{equation}
c = \sqrt{q_0^2 + q_1^2 + q_2^2 + q_3^2} = 1
\label{eq:c}
\end{equation}

\begin{equation}
\mathbf{A}_{[n \times 7]} = \left. \frac{\partial f}{\partial \mathbf{X}} \right|_{\mathbf{X}_0, \mathbf{V}_0},
\qquad
\mathbf{B}_{[n \times 3n]} = \left. \frac{\partial f}{\partial \mathbf{V}} \right|_{\mathbf{X}_0, \mathbf{V}_0},
\qquad
\mathbf{C}_{[1 \times 7]} = \left. \frac{\partial c}{\partial \mathbf{X}} \right|_{\mathbf{X}_0}.
\label{eq:ABC}
\end{equation}

$n$ is the number of observations and $\mathbf{V}$ is the correction of observations. $\mathbf{C}$ is the matrix composed of the derivatives of the constraint $c$ concerning $\mathbf{X}$. It is worth noting that when only facade-based correspondences are available, the structure of the matrix \( \mathbf{A} \) takes the following form: 

\begin{equation}
\mathbf{A}_{\text{facade only}} =
\frac{\partial f}{\partial \mathbf{X}} =
\begin{bmatrix}
\frac{\partial f_1}{\partial q_0} & \frac{\partial f_1}{\partial q_1} & \frac{\partial f_1}{\partial q_2} & \frac{\partial f_1}{\partial q_3} & \frac{\partial f_1}{\partial t_x} & \frac{\partial f_1}{\partial t_y} &  0 \\
\frac{\partial f_2}{\partial q_0} & \frac{\partial f_2}{\partial q_1} & \frac{\partial f_2}{\partial q_2} & \frac{\partial f_2}{\partial q_3} & \frac{\partial f_2}{\partial t_x} & \frac{\partial f_2}{\partial t_y} &  0 \\
\vdots & \vdots & \vdots & \vdots & \vdots & \vdots & \vdots \\
\end{bmatrix}.
\end{equation}

Due to the absence of ground-plane constraints, the rightmost column of \( \mathbf{A} \), corresponding to the vertical translation \( t_z \), consists entirely of zeros, resulting in a rank-deficient system in the vertical direction. This may result in an unsolvable or invalid solution.

The incorporation of the pseudo-plane correspondence introduces an additional ground-plane constraint. As two pseudo-planes are deliberately constructed with identical parameters and perfectly aligned geometry, this data-driven constraint introduces no residual, but provides an additional condition that reinforces the existing GHM system. The structure of the new matrix \( \mathbf{A} \) is converted to the following form:

\begin{equation}
\mathbf{A}_{\text{with pseudo-plane}} =
\frac{\partial f}{\partial \mathbf{X}} =
\begin{bmatrix}
\frac{\partial f_1}{\partial q_0} & \frac{\partial f_1}{\partial q_1} & \frac{\partial f_1}{\partial q_2} & \frac{\partial f_1}{\partial q_3} & \frac{\partial f_1}{\partial t_x} & \frac{\partial f_1}{\partial t_y} &  0 \\
\frac{\partial f_2}{\partial q_0} & \frac{\partial f_2}{\partial q_1} & \frac{\partial f_2}{\partial q_2} & \frac{\partial f_2}{\partial q_3} & \frac{\partial f_2}{\partial t_x} & \frac{\partial f_2}{\partial t_y} &  0 \\
\vdots & \vdots & \vdots & \vdots & \vdots & \vdots & \vdots \\
0 & 0 & 0 & 0 & 0 & 0 & 1 \\
\end{bmatrix}.
\end{equation}

As a result, 3D rotation and horizontal translation (\( t_x \) and \( t_y \)) can be accurately estimated, because the \( t_z \) currently obtained is always zero or very close to zero. The actual vertical translation \( t_z \) will be solved separately in the following Section \ref{subsec3.4}. 

The primary advantage of this strategy lies in its ability to perform reliable parameter estimation even in the absence of real ground-plane correspondences. More importantly, it effectively decouples the influence of low-quality observations--such as those derived from DTM data--from high-quality observations, preventing error propagation across components and maximizing the use of reliable facade-based correspondences. 

On the one hand, such a design remains adaptable to ground data of varying types, accuracies, and acquisition sources. This adaptability becomes even more evident in scenarios with sloped or uneven ground surfaces, where simply constraining or omitting the parameter in a given direction--such as \( t_z \) in the GHM--may still introduce additional errors. On the other hand, rather than omitting the estimation of \( t_z \) and solving only for five parameters, the pseudo-plane strategy offers greater flexibility and adaptability, making it readily extendable to other facade orientations. Overall, the pseudo-plane strategy delivers a balanced design that preserves estimation robustness while maintaining adaptability to diverse scenarios.

\subsection{Adaptable Vertical Translation Estimation}
\label{subsec3.4}

\begin{figure}[hbtp]
\centering
\includegraphics[width=1.0\linewidth]{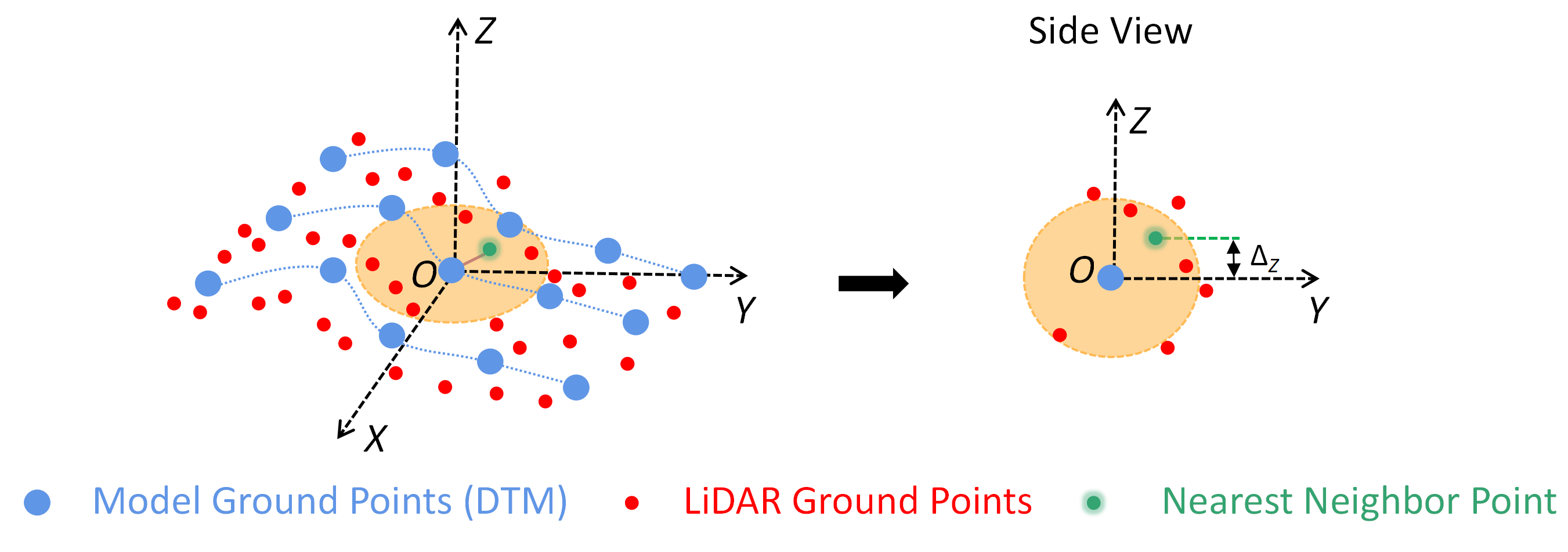}
\caption{Vertical translation estimation.}\label{height estimation}
\end{figure}

In this step, the goal is to estimate the remaining vertical translation \( t_z \). The transformation parameters obtained from the previous step are first applied to the LiDAR point clouds to achieve alignment in all directions except the vertical axis. Next, DTM data is utilized to establish point correspondences. Specifically, after denoising the transformed ground point clouds, each DTM point is used as a reference center, and its nearest neighbors within a certain radius in the $XOY$ plane are searched in the point clouds. As seen in Figure \ref{height estimation}, these local neighbors are then used to construct a set of vertical correspondences. The average deviation in the Z-coordinates across all correspondences is computed to obtain the final estimate of the vertical translation \( t_z \).

In this way, the vertical translation \( t_z \) can be easily estimated from DTM data while preserving a high degree of flexibility, benefiting from the aforementioned 2D-3D decoupled estimation strategy. For instance, if more accurate vertical information becomes available--such as GNSS observations, total station measurements, or other high-fidelity ground models--the DTM can be seamlessly substituted using the same method. This design greatly enhances the adaptability of the overall framework to heterogeneous data sources and ensures robustness under varying data availability conditions.

\section{Experiment and Results}
\label{sec4}

This part introduces the experiment and results, including data preparation in Section \ref{subsec4.1}, experiment design in Section \ref{subsec4.2}, evaluation metrics in Section \ref{subsec4.3}, and comparison results in Section \ref{subsec4.4}.

\subsection{Data Preparation}
\label{subsec4.1}

\begin{figure}[hbtp]
\centering
\includegraphics[width=1.0\linewidth]{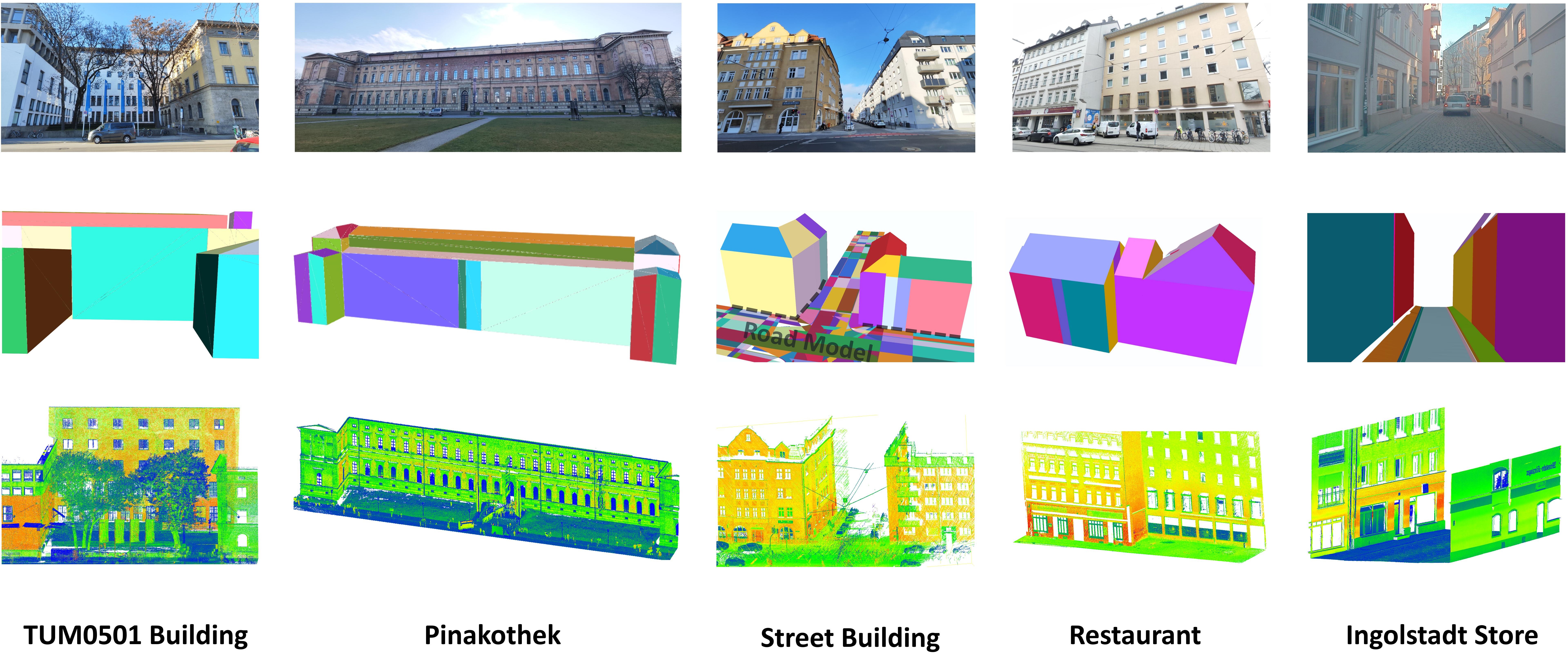}
\caption{Overview of five datasets. The upper part shows images of each building. LoD2 semantic models are in the middle, with planes colored for visual distinction. The lower part displays the corresponding point clouds, colored by intensity. For better visualization, the scale of each dataset in this figure is not directly comparable. Quantitative information is provided in Table \ref{tabDatasets}.}\label{FigDataset}
\end{figure}

To validate and evaluate the proposed L2M-Reg, four datasets from Munich (as shown in Figure \ref{FigDataset}), named as TUM0501 Building, Pinakothek, Street Building, and Restaurant, are collected, comprising LiDAR point clouds from TLS or MLS. Besides, the corresponding publicly accessible LoD2 models are obtained from the Bavarian Surveying Administration\footnote{\url{https://geodaten.bayern.de/opengeodata/}}. Additionally, one more dataset from another city, Ingolstadt, named Ingolstadt Store, is included in the experiment. These five datasets vary in point cloud size, location, building size, and data acquisition methods. It is worth noting that the two buildings, Restaurant and Ingolstadt Store, not only contain the wall plinths examined in this study, but also feature extensive glass elements on their exterior facades, including windows, street-facing glass doors, and glass display windows. In the Restaurant dataset, some of these glass surfaces are even covered with posters. All facades across the five datasets examined contain a certain degree of non-planar components, such as rainwater pipes, attached mailboxes, among others. As a result, the input point clouds used for plane extraction include not only the target planar regions but also several discrete non-planar structures.

In particular, due to the absence of ground data in the publicly accessible LoD2 models, ground information for the TUM0501 Building and Pinakothek datasets, which are part of the TUM2TWIN\footnote{\url{https://tum2t.win/}} \citep{wysocki2025}, is supplemented by publicly available DTM data with a grid width of 1 m from the Bavarian Surveying Administration. The Restaurant dataset also has the same DTM data. The Street Building and Ingolstadt Store datasets utilize additional georeferenced road models. Further details of the datasets are summarized in Table \ref{tabDatasets}.

\begin{table}[h]
\centering
\caption{Basic Information on Five Datasets}
\resizebox{\textwidth}{!}{
\begin{tabular}{cccccc}
\hline

                     & Number of Points & Scanning Area & Scanning System &  Ground Data Source & Location\\ \hline
 \textbf{TUM0501 Building} & 2,272,810 & $\approx$ 1,200 m$^{2}$ & Leica ScanStation P50 & DTM & Munich\\
 \textbf{Pinakothek} & 10,367,229 & $\approx$ 3,600 m$^{2}$ & Leica ScanStation P50 & DTM & Munich\\
 \textbf{Street Building} & 5,000,509 & $\approx$ 1,500 m$^{2}$ & Z+F FlexScan 22 & Road Model & Munich\\
 \textbf{Restaurant} & 24,116,910 & $\approx$ 800 m$^{2}$ & Z+F FlexScan 22 & DTM & Munich\\
 \textbf{Ingolstadt Store} & 7,477,684 & $\approx$ 60 m$^{2}$ & MoSES \citep{haigermoser2015} & Road Model & Ingolstadt\\
  \hline
  \end{tabular}
}
\label{tabDatasets}
\end{table}

\subsection{Experiment Design}
\label{subsec4.2}

In addition to the proposed L2M-Reg, five existing methods--GICP \citep{segal2009}, TriICP \citep{chetverikov2002}, KISS-ICP \citep{vizzo2023}, PLADE \citep{chen2019}, and Scantra \citep{Wujanz2018}--are also applied to the five datasets for performance evaluation and comparison. GICP, TriICP, and KISS-ICP are chosen as leading robust ICP-variants, known for their reliability in fine-registration tasks. PLADE and Scantra are included as the two leading plane-based registration methods. Since these registration methods are inherently designed for point clouds rather than parametric models, the LoD2 models from the five datasets were converted into point clouds with uniform density to facilitate registration using the five methods mentioned above.

L2M-Reg was implemented in C++ using Point Cloud Library (PCL)  \citep{rusu20113}. Standard implementations of GICP and TriICP in PCL (version 1.13.0) were utilized, while PLADE was implemented based on its open-source release\footnote{\url{https://github.com/chsl/PLADE}} \citep{chen2019}. Scantra was tested on the basis of the latest version 3.4 released in 2025\footnote{\url{https://www.technet-gmbh.com/en/products/scantra/scantra-news/scantra-release-34-static-kinematic-and-polar-are-ready-for-download/}}. KISS-ICP was reproduced using its open-source implementation \footnote{\url{ https://github.com/PRBonn/kiss-icp}}. All methods were executed under identical hardware conditions (Intel Xeon(R) W-2223 CPU@3.60GHz with 4 Cores and 64 GB of RAM).

\subsection{Evaluation Metrics}
\label{subsec4.3}

To quantitatively evaluate the registration accuracy, the M3C2 distance \citep{lague2013} between the registered LiDAR point clouds and the reference model is computed. Subsequently, a set of check points $p$ is uniformly selected from stable regions near building footprints, extending in various directions within each scene, to assess horizontal ($p\in H_i$) and vertical ($p\in V_i$) registration error. The average horizontal error $\mathit{Err}_{\mathrm{H}}$ and vertical error $\mathit{Err}_{\mathrm{V}}$ are defined by the average M3C2 distance at all check points, as calculated by

\begin{equation}
\mathit{Err}_{\mathrm{H}} = \frac{1}{n_H} \sum_{\substack{p \in H_i}} \delta_{\mathrm{M3C2}}(p),
\label{eq:acc_h}
\end{equation}

\begin{equation}
\mathit{Err}_{\mathrm{V}} = \frac{1}{n_V} \sum_{p \in V_i} \delta_{\mathrm{M3C2}}(p),
\label{eq:acc_v}
\end{equation}

where $n_H$ and $n_V$ denote the number of check points used for assessing horizontal and vertical error, respectively. For each check point $p$, $\delta_{\mathrm{M3C2}}(p)$ represents the M3C2 distance between $p$ and the reference model. Similarly, based on Eq.~\eqref{eq:acc_h} and Eq.~\eqref{eq:acc_v}, the standard deviation of M3C2 distances in the horizontal and vertical directions can be computed by

\begin{equation}
\mathit{Std}_{\mathrm{H}} = \sqrt{ \frac{1}{n_H - 1} \sum_{p \in H_i} \left( \delta_{\mathrm{M3C2}}(p) - \mathit{Err}_{\mathrm{H}} \right)^2 },
\label{eq:var_h}
\end{equation}

\begin{equation}
\mathit{Std}_{\mathrm{V}} = \sqrt{ \frac{1}{n_V - 1} \sum_{p \in V_i} \left( \delta_{\mathrm{M3C2}}(p) - \mathit{Err}_{\mathrm{V}} \right)^2 }.
\label{eq:var_v}
\end{equation}

These evaluation metrics are used for two primary reasons. First, the M3C2 distance is more robust to noise and non-uniform point spacing than the nearest Cloud-to-Cloud (C2C) distance, providing a more accurate measure of geometric consistency between point clouds and models. Second, sampling from geometrically stable regions (i.e., footprint area) in both horizontal and vertical directions reduces the influence of dynamic elements such as vegetation and pedestrians, thus improving the reliability of the accuracy assessment.

\subsection{Comparison Results}
\label{subsec4.4}

The following part systematically demonstrates the performance of L2M-Reg in terms of accuracy and efficiency through qualitative comparison in Section \ref{subsubsection4.4.1}, quantitative comparison in Section \ref{subsubsection4.4.2}, and efficiency analysis in Section \ref{subsubsection4.4.3}.

It is important to note that, unlike conventional evaluations based on the overall geometric closeness of the entire building, this study--motivated by the focus on model uncertainty--adopts the building footprint area as the primary basis for evaluation. This choice stems from the fact that, during LoD2 model generation, the near-ground footprint area is typically the most suited and accurately reflects the building’s true location, especially when there is a horizontal offset between plinth and facade. Accordingly, all subsequent evaluations in this study assess horizontal accuracy merely based on the distance between the aligned LiDAR point clouds and LoD2 models in the footprint area.

\subsubsection{Qualitative Comparison}
\label{subsubsection4.4.1}

\begin{figure}[hbtp]
\centering
\includegraphics[width=1\linewidth]{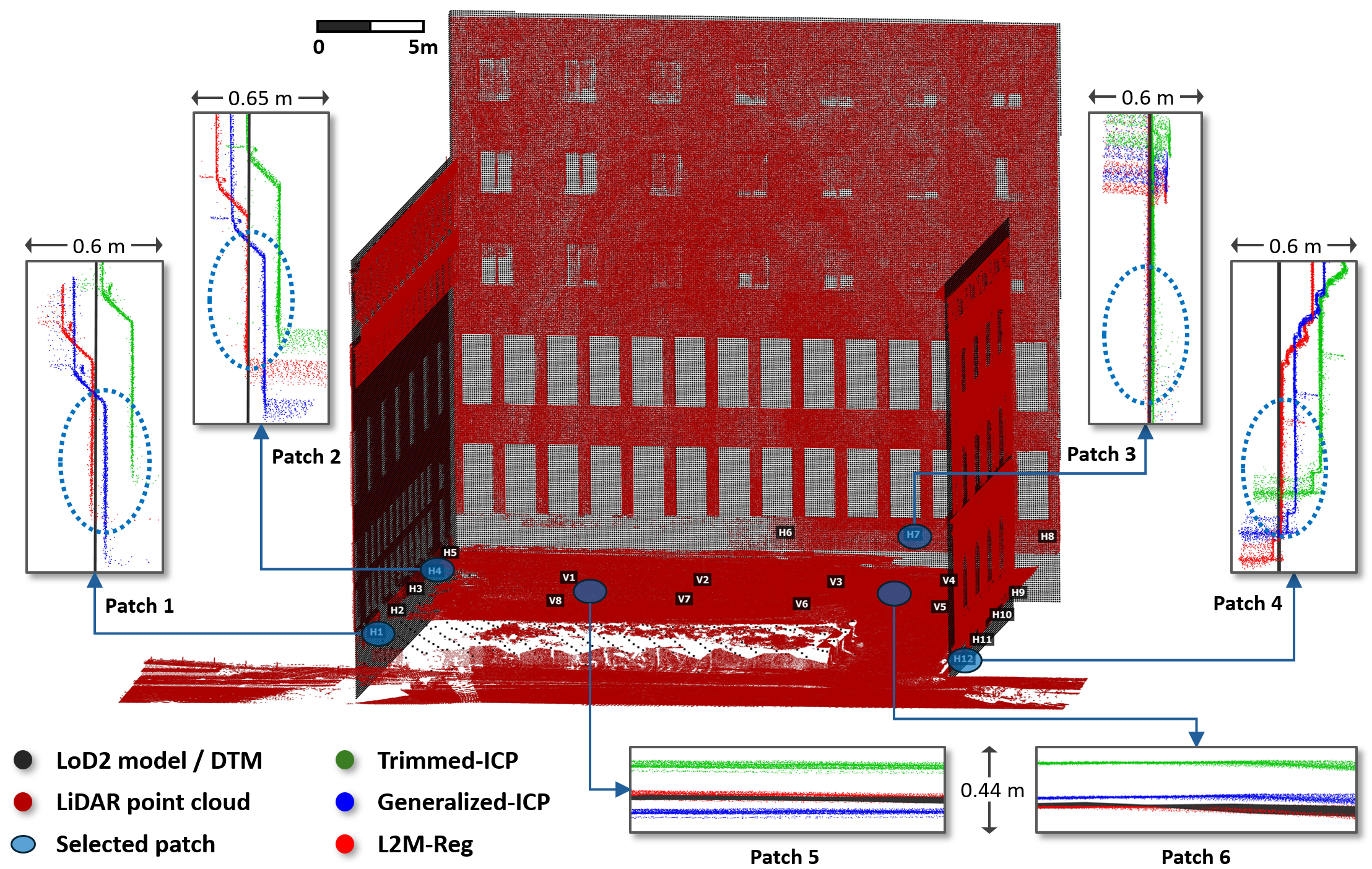}
\caption{Registration performance of different methods in the TUM0501 Building dataset.}
\label{qualitative0501}
\end{figure}

\begin{figure}[hbtp]
\centering
\includegraphics[width=1\linewidth]{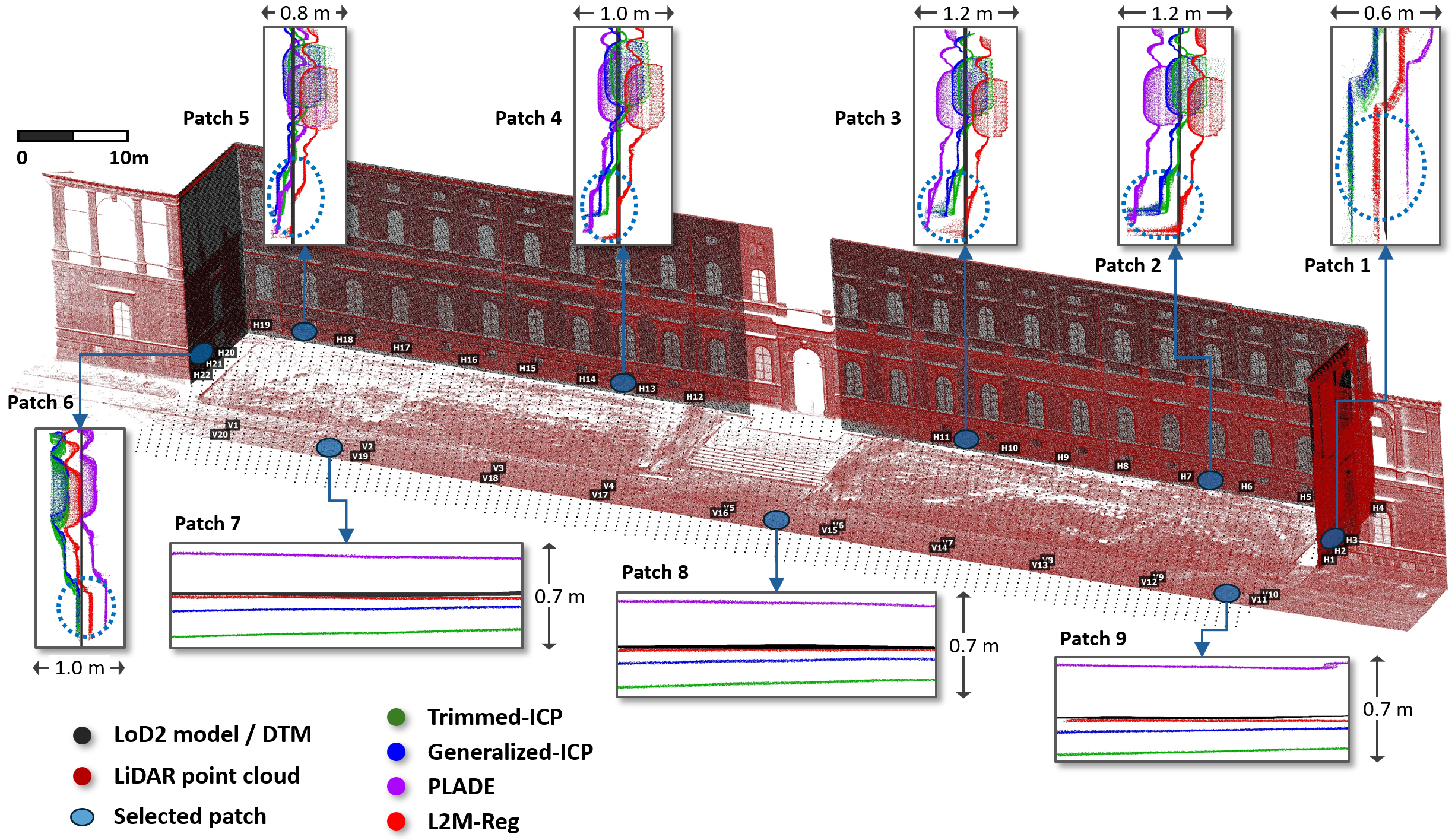}
\caption{Registration performance of different methods in the Pinakothek dataset.}
\label{qualitativePina}
\end{figure}

\begin{figure}[hbtp]
\centering
\includegraphics[width=1.0\linewidth]{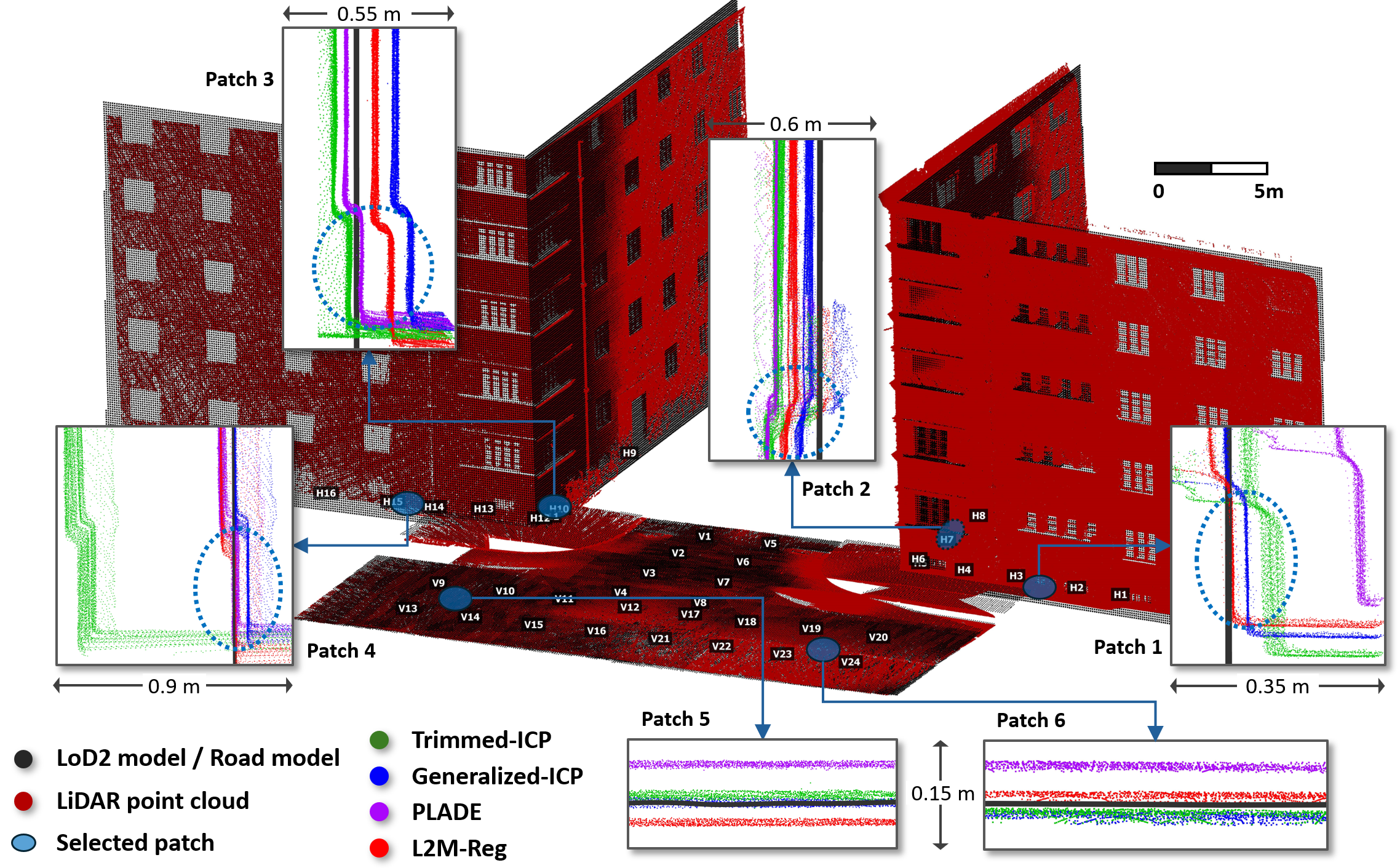}
\caption{Registration performance of different methods in the Street Building dataset. Patches 1 to 4 correspond to the registration results of the four facades from right to left. Due to viewpoint occlusion, Patch 2 is located on the second facade from the right.}
\label{qualitativeStreet}
\end{figure}

\begin{figure}[hbtp]
\centering
\includegraphics[width=1.0\linewidth]{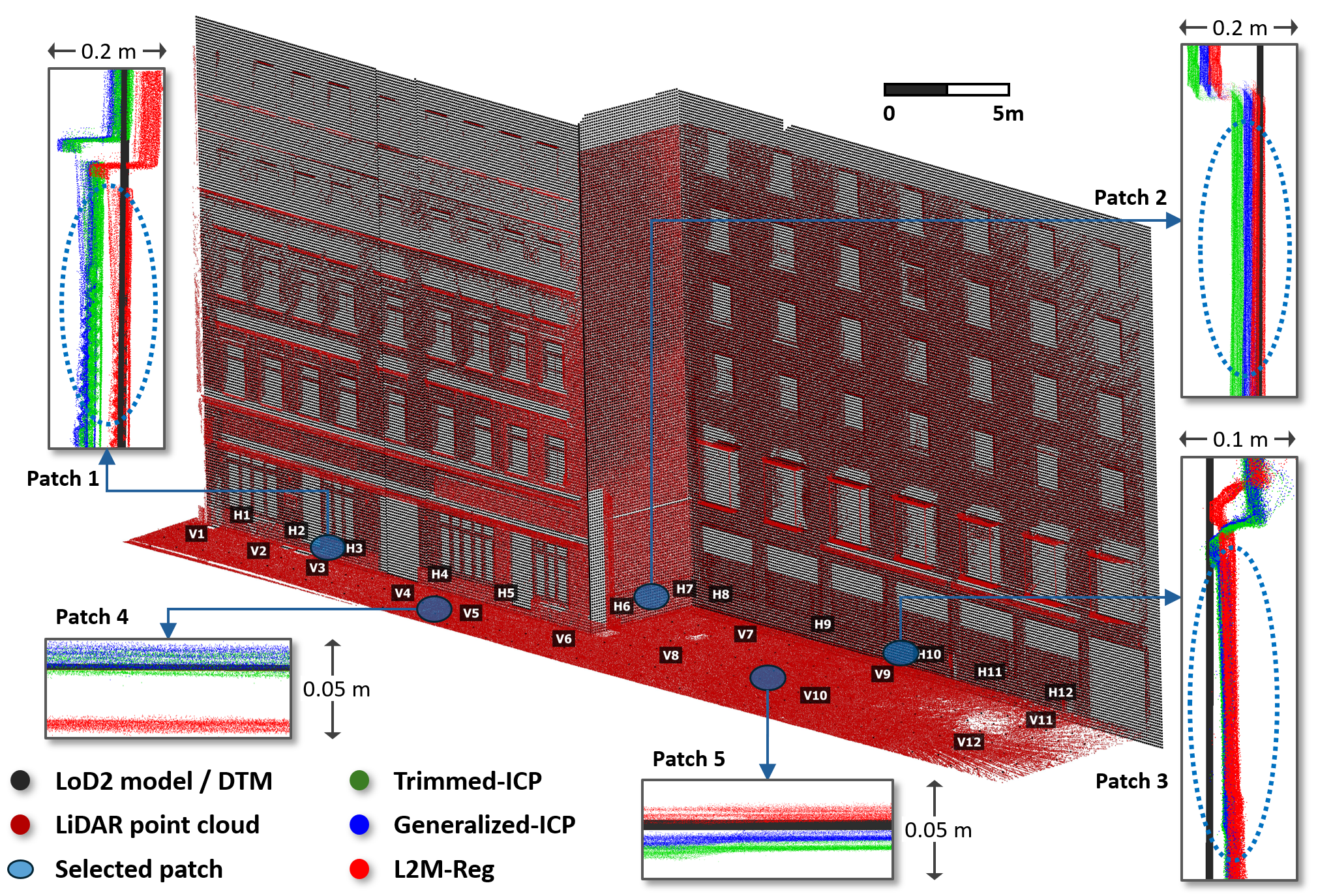}
\caption{Registration performance of different methods in the Restaurant dataset.}
\label{qualitativeRestaurant}
\end{figure}

\begin{figure}[hbtp]
\centering
\includegraphics[width=1.0\linewidth]{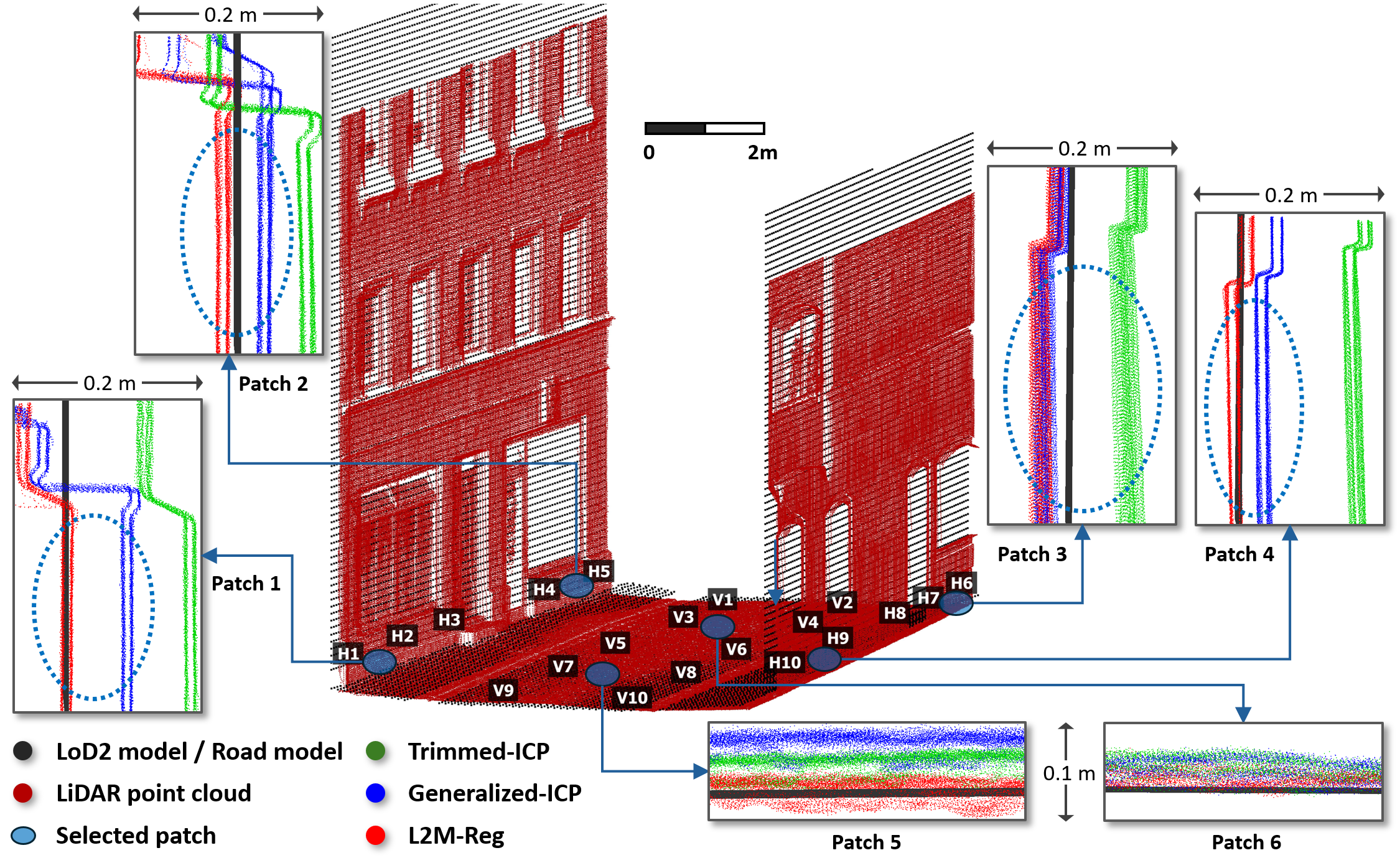}
\caption{Registration performance of different methods in the Ingolstadt Store dataset.}
\label{qualitativeIngo}
\end{figure}

Figures~\ref{qualitative0501}, \ref{qualitativePina}, \ref{qualitativeStreet}, \ref{qualitativeRestaurant}, and \ref{qualitativeIngo} show the registration results for the TUM0501, Pinakothek, Street Building, Restaurant, and Ingolstadt Store datasets, respectively. During testing, KISS-ICP, TriICP, GICP, and L2M-Reg consistently produced valid results for all datasets, whereas PLADE failed to do so for the TUM0501 Building, Restaurant, and Ingolstadt Store. Scantra exhibited suboptimal performance across all datasets. Possible reasons for these shortcomings are discussed in Section \ref{sec5}.

In addition, due to the high similarity in visual detail between KISS-ICP and other effective methods, visual distinction is very difficult. To improve clarity in visualization, results of KISS-ICP are not presented in the qualitative comparison of Figures \ref{qualitative0501} $\sim$ \ref{qualitativeIngo}. The complete results of all methods can be seen in the subsequent quantitative evaluation.

As shown in Figure \ref{qualitative0501}, the TUM0501 Building dataset yields registration results using TriICP, GICP, and L2M-Reg. The results of PLADE are not included, as it can not produce valid results. In the horizontal direction, particularly within the building footprint areas outlined by dashed circles, L2M-Reg achieves the highest consistency compared to TriICP and GICP. In the vertical direction, it also shows superior alignment with the ground model, further highlighting its advantage in registration accuracy.

Figure \ref{qualitativePina} shows registration results for the Pinakothek dataset. In the horizontal direction, L2M-Reg consistently demonstrates higher accuracy within the footprint area, as illustrated in Patch 1 to Patch 5 (outlined by dashed circles). Notably, in Patch 6, L2M-Reg exhibits a visible deviation from the reference model compared to the two ICP-based methods. This discrepancy is primarily attributed to scale inaccuracies in the LoD2 model, resulting in dimensional mismatches between the model and the point cloud. Despite this, L2M-Reg still achieves the globally optimal registration, as evidenced by Patch 1 being parallel to Patch 6, where L2M-Reg again provides a better global fit. In the vertical direction, as shown in Patch 7 to Patch 9, L2M-Reg demonstrates the best alignment.

In the Street Building dataset, shown in Figure \ref{qualitativeStreet}, L2M-Reg consistently delivers the most accurate alignment in the horizontal direction, particularly in Patch 1 and Patch 4. Similar slight dimensional mismatches between the models and the point clouds are observed in Patch 2 and Patch 3, as seen in the Pinakothek dataset. These mismatches are again attributed to scale inaccuracies introduced in the original models. Nevertheless, L2M-Reg achieves a globally optimal result, with the registered point clouds (in red) centrally aligned between the two corresponding facades of the LoD2 models in Patch 2 and Patch 3.

In the vertical direction (e.g., Patch 6), L2M-Reg also achieves the best alignment in most regions. However, in certain areas like Patch 5, it does not produce optimal results. This results from the road being originally described according to the OpenDRIVE standard \citep{asamOpenDRIVEV181User2024,kutschTUMDOTMUCData2024}, which models individual lanes with overlapping surface geometries and continuous curvature. The parametric surface representations are then sampled to explicit planar surface geometries as part of the conversion to CityGML 3.0 \citep{schwabSpatiosemanticRoadSpace2020}. Although unintended overlapping surface geometries introduce minor instability in the evaluation of vertical accuracy, this demonstrates that L2M-Reg can also directly incorporate further object classes of the model, such as the road surface, into the vertical alignment process. Since CityGML 3.0 enables a comprehensive and redundancy-free representation of the road space, such ambiguities are not introduced in native mapping processes.

As shown in Figure \ref{qualitativeRestaurant}, the Restaurant dataset yields registration results using TriICP, GICP, and L2M-Reg. PLADE is not able to give valid results in this case. In the horizontal direction, particularly within the building footprint areas outlined by dashed circles in Patch 1 to Patch 3, L2M-Reg achieves the highest consistency. In the vertical direction, it also shows superior alignment, as illustrated in Patch 5. However, in Patch 4, its performance falls slightly short of the optimal method.
 
Figure \ref{qualitativeIngo} shows the registration results for the Ingolstadt Store dataset. In the horizontal direction, L2M-Reg consistently demonstrates the best alignment within the footprint area, as illustrated in Patch 1 to Patch 4 (outlined by dashed circles). In the vertical direction, as shown in Patch 5 to Patch 6, L2M-Reg also demonstrates the best alignment between point clouds and road model.

In general, the qualitative comparison results indicate that L2M-Reg outperforms the other methods in both horizontal and vertical consistency in all datasets. More detailed quantitative evaluation results are presented in Section \ref{subsubsection4.4.2}.

\subsubsection{Quantitative Comparison}
\label{subsubsection4.4.2}

In this section, the aforementioned evaluation metrics are used to quantitatively compare the five valid registration methods (KISS-ICP, TriICP, GICP, PLADE, and L2M-Reg) based on the check points manually selected from stable areas. Table \ref{tabCheckPoint} shows the number of check points used in each dataset. In the Street Building dataset, all ground points in the central areas of the road were used for vertical error evaluation, considering local layering issues in the road model.

\begin{table}[h]
\centering
\caption{The Number of Check Points Used in Five Datasets}
\label{tabCheckPoint}
\resizebox{0.8\textwidth}{!}{
\begin{tabular}{ccc}
\hline

        & Horizontal check points   & Vertical check points \\ \hline
  \textbf{TUM0501 Building}  & 12 & 8 \\
  \textbf{Pinakothek}  & 22  & 20 \\
  \textbf{Street Building} & 16 & 620 \\
  \textbf{Restaurant} & 12 & 12 \\
  \textbf{Ingolstadt Store} & 10 & 10 \\
  \hline
  \end{tabular}
  }
\vspace{0.5em}
\end{table}

Table \ref{tabAccuracy} reports the average registration error in both horizontal and vertical directions. L2M-Reg achieves the best performance on all datasets, which is consistent with the qualitative results discussed before. There are only two exceptions regarding the vertical error of the Street Building and Restaurant datasets, where L2M-Reg performs slightly worse. This minor discrepancy is primarily attributed to the reference ground data, which is derived from an additional road model rather than a standard DTM for the Street Building. Minor layering artifacts in this road model introduce local inconsistencies that bias the vertical error calculation. Nevertheless, the resulting loss (around 0.5 cm) is negligible for most practical purposes.
\begin{table}[h]
\centering
\normalsize
\caption{Average Error Comparison on Five Datasets (cm)}
\label{tabAccuracy}
\begin{tabular}{ccccccc}
\hline
& & KISS-ICP & TriICP & GICP & PLADE & L2M-Reg (Ours) \\
\hline
\multirow{2}{*}{\textbf{TUM0501 Building}} 
& $\mathit{Err}_{\mathrm{H}}$ & 8.68 & 14.07 & 5.75 & \textbf{--} & \textbf{0.98} \\
& $\mathit{Err}_{\mathrm{V}}$ & 6.74 & 24.44 & 6.24 & \textbf{--} & \textbf{1.20} \\
\cdashline{2-7}
\multirow{2}{*}{\textbf{Pinakothek}} 
& $\mathit{Err}_{\mathrm{H}}$ & 26.75 & 12.23 & 18.13 & 32.36 & \textbf{3.01} \\
& $\mathit{Err}_{\mathrm{V}}$ & 2.82 & 24.86 & 9.20  & 31.13 & \textbf{2.03} \\
\cdashline{2-7}
\multirow{2}{*}{\textbf{Street Building}} 
& $\mathit{Err}_{\mathrm{H}}$ & 9.18 & 24.31 & 6.68 & 11.77 & \textbf{4.87} \\
& $\mathit{Err}_{\mathrm{V}}$ & 2.80 & 2.89  & \textbf{1.89}  & 6.71 & 2.30 \\
\cdashline{2-7}
\multirow{2}{*}{\textbf{Restaurant}} 
& $\mathit{Err}_{\mathrm{H}}$ & 9.22 & 5.75 & 5.90 & \textbf{--} & \textbf{5.29} \\
& $\mathit{Err}_{\mathrm{V}}$ & \textbf{1.36} & 1.92 & 1.64 & \textbf{--} & 1.91 \\
\cdashline{2-7}
\multirow{2}{*}{\textbf{Ingolstadt Store}} 
& $\mathit{Err}_{\mathrm{H}}$ & 11.97 & 13.16 & 4.44 & \textbf{--} & \textbf{1.23} \\
& $\mathit{Err}_{\mathrm{V}}$ & 4.16 & 1.17 & 1.24 & \textbf{--} & \textbf{0.78} \\
\hline
\end{tabular}
\vspace{0.5em}
\noindent\makebox[\textwidth][c]{\footnotesize Note: "--'' indicates that the method failed to produce valid results; Bold values indicate the best results.}
\end{table}

Table \ref{tabVariances} presents the standard deviation of the M3C2 distance across all check points. L2M-Reg consistently achieves the lowest standard deviation across all datasets except the Restaurant, indicating its effectiveness in producing globally optimal and stable registration results. A similar issue can be observed again in the vertical direction of the Street Building dataset. As previously noted, this slight inconsistency primarily stems from the characteristics of the road model. While L2M-Reg performs slightly worse than the best-performing method on the Restaurant dataset, the differences are all below 0.5 cm and are therefore considered acceptable.

\begin{table}[h]
\centering
\normalsize
\caption{Standard Deviation Comparison on Five Datasets (cm)}
\label{tabVariances}
\begin{tabular}{ccccccc}
\hline
& & KISS-ICP& TriICP & GICP & PLADE & L2M-Reg (Ours) \\
\hline
\multirow{2}{*}{\textbf{TUM0501 Building}} 
& $\mathit{Std}_{\mathrm{H}}$ & 3.44 & 7.46 & 3.39 & \textbf{--} & \textbf{0.81} \\
& $\mathit{Std}_{\mathrm{V}}$ & 5.08 & 3.24 & 3.71 & \textbf{--} & \textbf{0.72} \\
\cdashline{2-7}
\multirow{2}{*}{\textbf{Pinakothek}} 
& $\mathit{Std}_{\mathrm{H}}$ & 12.17 & 5.15 & 6.83 & 17.09 & \textbf{3.91} \\
& $\mathit{Std}_{\mathrm{V}}$ & 1.65 & 2.22 & 1.32 & 4.16 & \textbf{0.72} \\
\cdashline{2-7}
\multirow{2}{*}{\textbf{Street Building}} 
& $\mathit{Std}_{\mathrm{H}}$ & 8.51 & 22.3 & 6.06 & 9.99 & \textbf{6.05} \\
& $\mathit{Std}_{\mathrm{V}}$ & 2.05 & 2.19 & \textbf{1.37}  & 3.28 & 1.50 \\
\cdashline{2-7}
\multirow{2}{*}{\textbf{Restaurant}} 
& $\mathit{Std}_{\mathrm{H}}$ & 6.92 & 4.63 & \textbf{4.62} & \textbf{--} & 4.94 \\
& $\mathit{Std}_{\mathrm{V}}$ & 0.96 & 1.06 & \textbf{0.94} & \textbf{--} & 1.33 \\
\cdashline{2-7}
\multirow{2}{*}{\textbf{Ingolstadt Store}} 
& $\mathit{Std}_{\mathrm{H}}$ & 3.24 & 4.17 & 2.57 & \textbf{--} & \textbf{1.21} \\
& $\mathit{Std}_{\mathrm{V}}$ & 1.51 & 0.76 & 0.78 & \textbf{--} & \textbf{0.64} \\
\hline
\end{tabular}
\vspace{0.5em}
\noindent\makebox[\textwidth][c]{\footnotesize Note: ``--'' indicates that the method failed to produce valid results; Bold values indicate the best results.}
\end{table}

In summary, both qualitative and quantitative comparisons demonstrate that L2M-Reg achieves superior performance, as reflected in lower average error and standard deviations. For the Street Building and Restaurant datasets, although it does not exhibit a clear advantage, its performance remains fully comparable to that of the best-performing method. Overall, these comparisons prove the effectiveness of L2M-Reg and highlight its leading performance across diverse building scenarios.

\subsubsection{Efficiency Analysis}
\label{subsubsection4.4.3}

Table \ref{table runtime} summarizes the running times of L2M-Reg in comparison to KISS-ICP, GICP, TriICP, and PLADE on five datasets. All methods are evaluated under identical hardware conditions, with a maximum iteration limit of 100. Scantra is excluded from this comparison as it failed to yield valid results.

The results in Table \ref{table runtime} demonstrate that L2M-Reg exhibits promising computational efficiency, particularly on the Pinakothek and Restaurant datasets, which contain larger numbers of points. The primary reason is that GICP, TriICP, and PLADE operate on the entire point clouds, leading to computational times scaling linearly with the number of points. In contrast, L2M-Reg significantly reduces the computing time by first fixing correspondence relationships within data association and selectively processing points from the building plinth. Moreover, as a plane-based method, its complexity is less sensitive to point density, leading to notable improvements in runtime performance. 

Additionally, compared to KISS-ICP, L2M-Reg achieves superior accuracy as shown in Table \ref{tabAccuracy}, but loses its lead in computational efficiency. This is because KISS-ICP employs very aggressive downsampling in its design to accelerate processing. Such a design primarily serves real-time demanding scenarios like autonomous driving and robotics, thereby sacrificing certain accuracy for improved efficiency.

\begin{table}[h]
\centering
    \caption{Comparison of Computational Efficiency}
\resizebox{0.9\textwidth}{!}{
\begin{tabular}{cccccc}
\hline

                         & GICP& TriICP & KISS-ICP & PLADE& L2M-Reg (Ours) \\ \hline
\textbf{TUM0501 Building} & \underline{30.4s} & 38.5s & \textbf{5.7s} & 81.2s & 38.0s \\
\textbf{Pinakothek} & 329.5s & 356.1s & \textbf{18.5s} & 182.4s & \underline{135.9s} \\
\textbf{Street Building} & 328.6s & 120.2s & \textbf{10.8s} & 140.9s & \underline{52.7s} \\
\textbf{Restaurant} & 1221.0s & 328.5s & \textbf{31.8s} & 167.5s & \underline{88.6s} \\
\textbf{Ingolstadt Store} & 258.0s & 24.5s & \textbf{11.2s} & 80.5s & \underline{22.0s}\\ 
  \hline
  \end{tabular}
}
\label{table runtime}
\noindent\makebox[\textwidth][c]{\footnotesize Note: Best results are highlighted in bold, and the second best results are underlined.}
\end{table}

\section{Discussion}
\label{sec5}

This part further discusses the advantages of L2M-Reg in Section \ref{subsec5.1}, comparison among three plane-based methods in Section \ref{subsec5.2}, and reasons for using the road model in some datasets in Section \ref{subsec5.3}. Moreover, Section \ref{subsec5.4}, \ref{subsec5.5}, and \ref{subsec5.6} give further information regarding the ablation study of GC check, robustness to different initial poses, and generalizability and prospective practical applications of L2M-Reg, respectively, providing a deeper understanding of its performance and applicability. Current limitations of this study are discussed in Section \ref{subsec5.7}.

\subsection{Advantages of Using Existing Model Semantics and Pseudo-plane Constraint}
\label{subsec5.1}

Compared to existing ICP-based (e.g., KISS-ICP, GICP, and TriICP) and plane-based (e.g., PLADE and Scantra) registration methods that exclusively process point clouds generated from LoD2 models, L2M-Reg uniquely leverages built-in semantic information in LoD2 models. Using these semantic attributes and distinct plane identifiers, facade elements can be efficiently and accurately extracted, enabling precise plane correspondence establishment. This strategy bypasses the conventional two‑step workflow of feature extraction followed by correspondence matching. Instead, it directly streamlines correspondence identification by leveraging the existing semantic information in LoD2 models, thus improving the registration robustness and efficiency.

As one of the key innovations of L2M-Reg, introducing the pseudo-plane constraint offers three distinct advantages. First, it enables 2D-3D decoupled parameter estimation, effectively preventing low-quality elevation data from influencing the estimation of other parameters. Second, it is formulated within a clear and interpretable mathematical framework under the GHM and can be implemented with minimal complexity. Third, compared to the alternative of manually fixing \( t_z \) to zero in GHM to isolate vertical errors, the pseudo-plane strategy provides greater flexibility and adaptability, allowing for extension to other facade orientations. Moreover, in scenarios involving sloped or uneven ground surfaces, directly constraining \( t_z \) in GHM may introduce additional errors, further highlighting the advantages of the pseudo-plane strategy.

\subsection{Interpretation of Performance Differences among Three Plane-based Methods}
\label{subsec5.2}

As plane-based registration methods, L2M-Reg, along with PLADE and Scantra, all utilize plane features to estimate the optimal transformation. However, their performance differs significantly in the context of building-level registration tasks, mainly attributable to their implementation.

The design of PLADE and Scantra lies in extracting a large number of planes prior to establishing correspondences using geometric descriptors. While effective in general urban scenes, this strategy proves unsuitable for individual buildings, which typically contain a limited number of planes with diverse orientations. In addition, their correspondence establishment heavily relies on rich geometric structures and distinctive features. This makes them less effective when applied to geometrically simplified or feature-sparse LoD2 models of individual buildings, accounting for their performance in the evaluated scenarios.

\subsection{Explanation of Using Road Model}
\label{subsec5.3}

In the Street Building and Ingolstadt Store datasets, two local road models produced by different agencies are used instead of the DTM to provide constraints in the vertical direction. This choice is motivated by two main factors. First, increasingly detailed ground and road models, such as high-definition maps for autonomous driving, are becoming available for local areas. Although their spatial coverage is currently limited, these datasets often offer higher accuracy than DTM derived from ALS and can serve as more reliable sources of absolute elevation. Second, the effectiveness of L2M-Reg when using DTM data has already been demonstrated based on the TUM0501 Building, Pinakothek, and Restaurant datasets. Owing to the 2D-3D decoupled estimation strategy introduced in this study, the estimation of vertical translation parameters does not affect other transformation parameters. Therefore, using the road model in the other two datasets further illustrates the adaptability of L2M-Reg.

While a slight difference in vertical accuracy is observed between L2M-Reg and the best-performing baseline (GICP) in the quantitative results (see Tables \ref{tabAccuracy} and \ref{tabVariances}), this discrepancy is primarily attributed to redundant layers within the used road model in the Street Building dataset, which distort the accuracy evaluation. When higher-quality road data are available, such as in the Ingolstadt Store dataset, L2M-Reg can readily adapt to these data sources and provide accurate estimations. This also demonstrates that L2M-Reg can be applied to data generated by different agencies across various cities.

\subsection{Ablation Study on the Geometric Consistency Check}
\label{subsec5.4}

In real-world LiDAR acquisition and processing, point clouds may exhibit layering effects caused by trajectory estimation errors in MLS or multi-station registration errors in TLS. This effect is particularly evident on building facades and can potentially affect plane extraction. When facade point clouds contain multiple layers, conventional RANSAC-based methods tend to extract planes from a single dominant layer and discard the remaining data by default, leading to incomplete or biased plane estimates.

To address this issue, the Geometric Consistency (GC) check is incorporated to improve the robustness of the proposed plane extraction algorithm. It introduces an additional angle-constrained check that enables the reuse of compatible points from adjacent layers. This design preserves the inherent noise robustness of RANSAC while explicitly mitigating performance degradation caused by point cloud layering, thereby enhancing the reliability of the extracted planes.

To further discuss the effectiveness of the GC check, a wall from the TUM0501 Building was selected for controlled analysis. Since L2M-Reg is a plane-based solution, the accuracy of the final result is actually tied to the accuracy of the estimated plane normals. Therefore, different levels of point cloud layering and misalignment were simulated on the original data. Five study cases were designed with progressively increasing inter-layer misalignment angles of $0.5^\circ$, $1.0^\circ$, $1.5^\circ$, $2.0^\circ$, and $2.5^\circ$.

\begin{figure}[hbtp]
\centering
\includegraphics[width=1.0\linewidth]{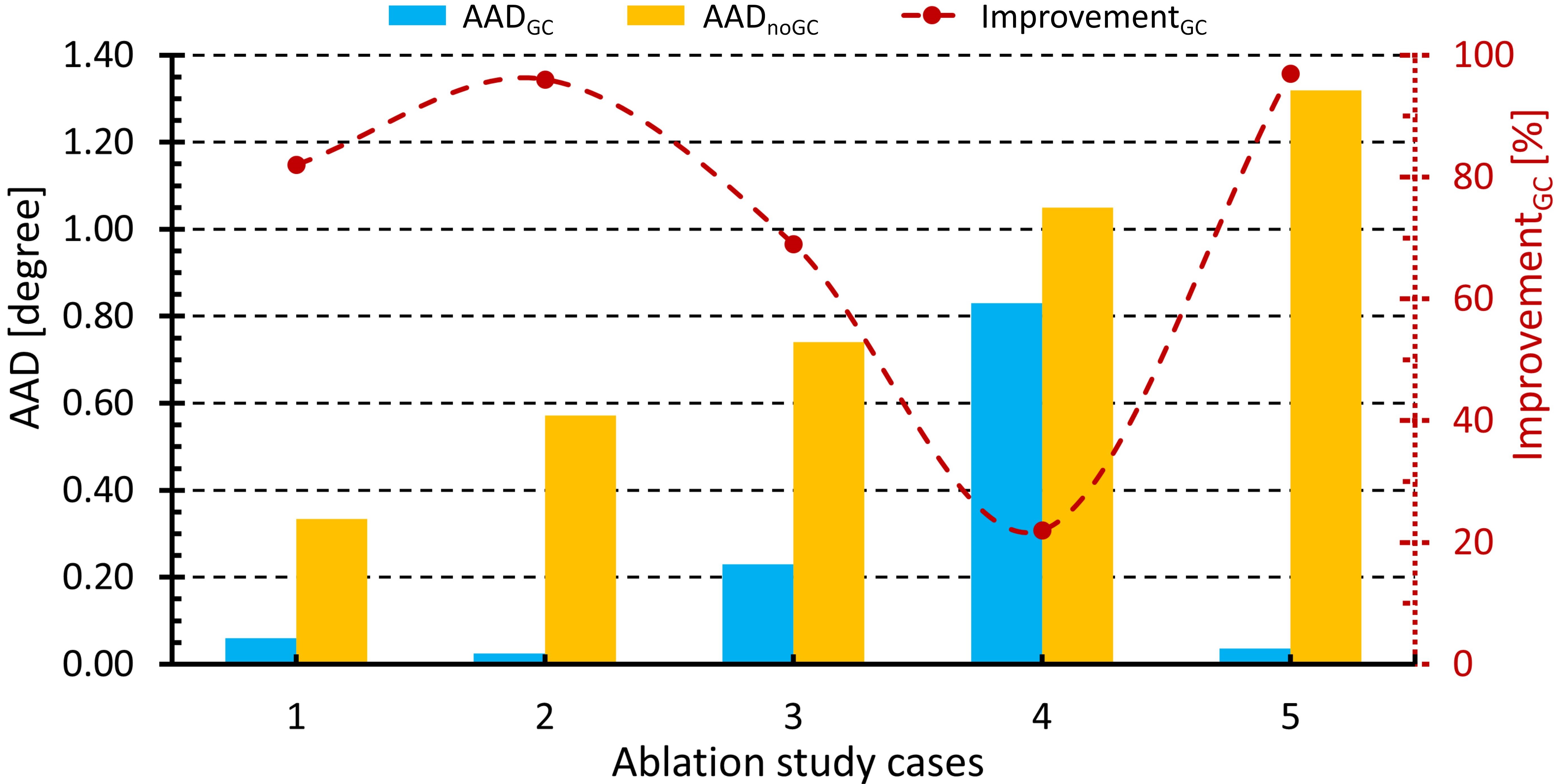}
\caption{Absolute Angular Deviation (AAD) with and without GC check, and the corresponding improvement benefiting from GC check in TUM0501 Building. Cases 1 to 5 correspond to inter-layer misalignment angles of $0.5^\circ$, $1.0^\circ$, $1.5^\circ$, $2.0^\circ$, and $2.5^\circ$, respectively.}
\label{ablation study}
\end{figure}

The results shown in Figure \ref{ablation study}, compare the Absolute Angular Deviation (AAD) of the extracted plane normals from the reference values with and without the GC check, abbreviated as $\mathit{AAD_{GC}}$ and $\mathit{AAD_{noGC}}$, respectively. In addition, the corresponding improvement percentage benefiting from the GC check is defined in Eq. \eqref{eq:improvement}:

\begin{equation}
\mathit{Improvement_{GC}} = \left( 1 - \frac{\mathit{AAD_{GC}}}{\mathit{AAD_{noGC}}} \right) \times 100\%.
\label{eq:improvement}
\end{equation}

Although the improvement varies across cases, it typically ranges from 20\% to 90\%, demonstrating the substantial contribution of the GC check to the accuracy of plane extraction under layered point cloud conditions.

\subsection{Robustness to Different Initial Poses from Coarse Registration}
\label{subsec5.5}

Additional experiments were conducted to examine the convergence behavior of L2M-Reg under varying levels of coarse registration errors. The TUM0501 Building was used as a representative example, and different rotations and translations were imposed on the original data to simulate increasing degrees of coarse registration errors.

\begin{figure}[hbtp]
\centering
\includegraphics[width=1.0\linewidth]{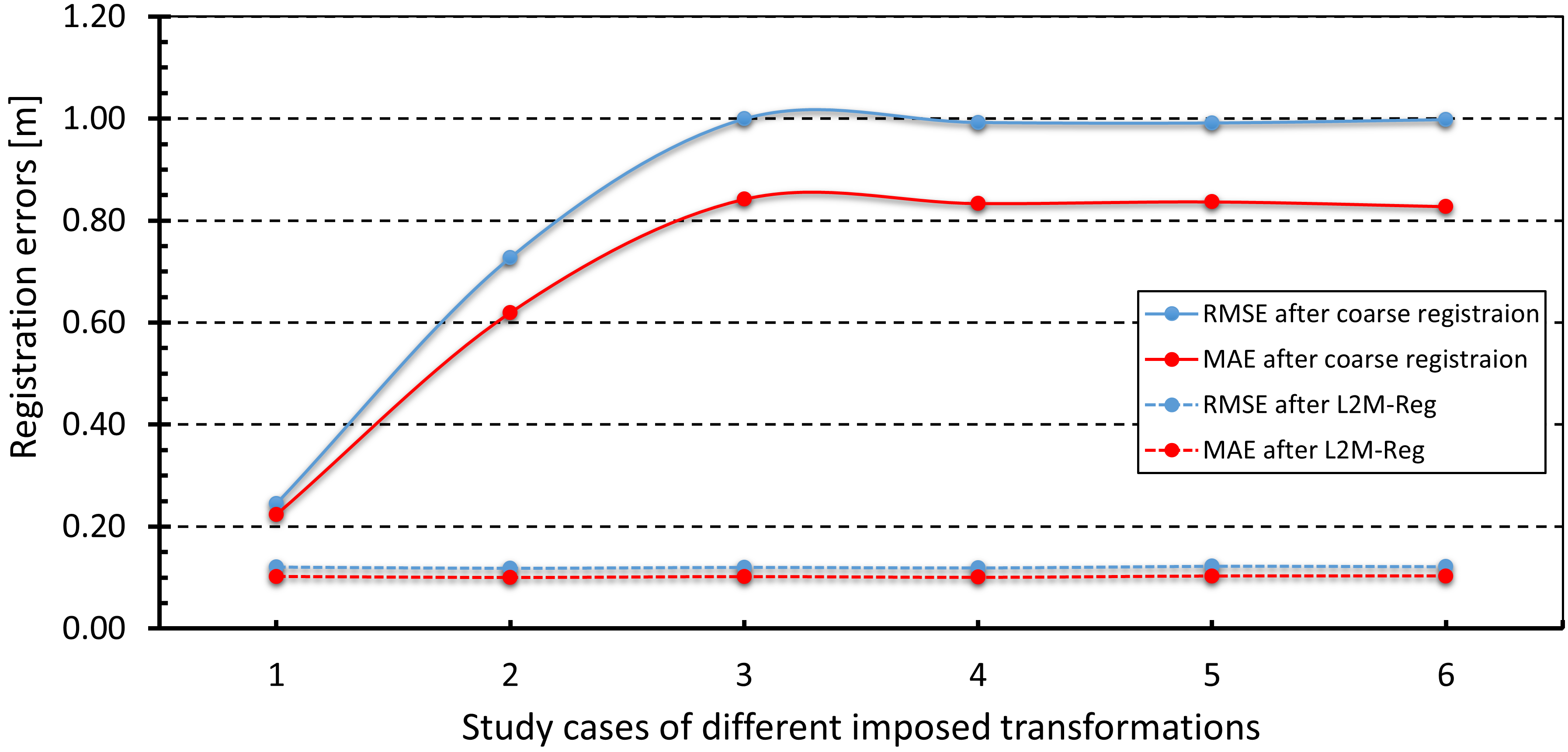}
\caption{Fine registration results from L2M-Reg under different coarse registration errors. The study case indices from 1 to 6 correspond to the imposed transformations introduced in Table \ref{table coarse registration}.}
\label{Coarse registration}
\end{figure}

In total, six study cases were designed. Case 1 serves as the baseline, while Cases 2 through 6 apply progressively larger transformations, as summarized in Table \ref{table coarse registration}. The magnitude of the simulated coarse registration errors increases across cases, as quantified by Root Mean Squared Error (RMSE) and Mean Absolute Error (MAE). The registration results obtained by L2M-Reg for all cases are shown in Figure \ref{Coarse registration}. The results indicate that L2M-Reg maintains robust performance and consistently achieves stable fine registration results, even when the RMSE or MAE after coarse registration approaches 1 m. In practice, coarse registration within a 1 m range is readily achievable using existing methods \citep{diakite2020, sheik2022a, sheik2022b}. Therefore, although L2M-Reg requires an initial coarse registration, it remains applicable to most practical scenarios.

\begin{table}[h]
\centering
    \caption{Different Imposed Transformations on Original Pose}
\resizebox{0.8\textwidth}{!}{
\begin{tabular}{ccl}
\hline

  & Study Case & Imposed Transformations \\ \hline
\multirow{6}{*}{\textbf{TUM0501 Building}} & 1 & Original \\
                                  & 2 & Original + \( R_z (5^\circ \)) \\
                                  & 3 & Original + \( R_z (10^\circ \)) \\
                                  & 4 & Original + \( R_z (15^\circ \)) \\
                                  & 5 & Original + \( R_z (15^\circ \)) +  \(t_z (0.5 m \)) \\ 
                                  & 6 & Original + \( R_z (20^\circ \)) +  \(t_z (1.0 m \)) \\ 
  \hline
  \end{tabular}
}
\label{table coarse registration}
\end{table}

Furthermore, to examine the convergence boundary of L2M-Reg under varying coarse registration errors more comprehensively, 12 additional test cases were constructed based on Case 6 in Table \ref{table coarse registration} and Figure \ref{Coarse registration}. All tests used the TUM0501 Building dataset. Larger imposed transformations were applied to simulate more severe coarse registration errors. In addition, multiple initial parameter settings within the GHM were considered. For each level of coarse registration error, four sets of initial parameters were randomly assigned for realistic processing conditions. The results, summarized in Table \ref{table coarse registration impact}, show that when the coarse registration RMSE is approximately 1 m, L2M-Reg consistently achieves high accuracy regardless of initialization. Similar behavior appears as the RMSE approaches 2 m (1.90 m), with no noticeable loss in convergence accuracy. However, when the initial RMSE increases to around 2.5 m, L2M-Reg's performance degrades. The average RMSE rises from 0.15 m to 0.21 m.

As an empirical rule of thumb derived from these experiments, L2M-Reg produces reliable results when the initial RMSE is within 2 m. Although convergence may still occur under larger errors, registration accuracy degrades and should be carefully considered in practice. This empirical rule of thumb is also influenced by other factors such as building size, point cloud quality, and structural complexity, and should therefore be adapted to specific scenarios. A more comprehensive investigation of coarse registration and its associated error characteristics is an important direction for future work but falls beyond the scope of this study. Nonetheless, the experiments demonstrate that L2M-Reg consistently delivers reliable performance when the initial RMSE is below 1 m, a level readily achievable by many existing methods.

\begin{table}[h]
\centering
    \caption{Impact of Different Coarse Registration Errors on L2M-Reg Performance}
\resizebox{1.0\textwidth}{!}{
\begin{tabular}{ccccccc}
\hline

\multirow{2}{*}{Dataset} & ImposedTrans & \multicolumn{2}{c}{CoarseRegErr before L2M-Reg} & InitialPara in GHM & \multicolumn{2}{c}{MeanRegErr after L2M-Reg}\\ 
\cdashline{2-7}
                         & \((R_x,R_y,R_z,t_x,t_y,t_z)\) & RMSE[m] & MAE[m] & \((R_x,R_y,R_z,t_x,t_y,t_z)\) & RMSE[m] & MAE[m] \\ 
\hline
\multirow{12}{*}{\textbf{TUM0501 Building}} & \multirow{4}{*}{\((0,0,20,0,0,1)\)} & \multirow{4}{*}{0.99} & \multirow{4}{*}{0.82} & \((0,0,0,0,0,0)\) & \multirow{4}{*}{0.14} & \multirow{4}{*}{0.12} \\
                                            &                                     &                       &                       & \((0,0,5,0,0,0.5)\) &                     &                       \\
                                            &                                     &                       &                       & \((0,0,10,0,0,0.8)\) &                    &                       \\
                                            &                                     &                       &                       & \((10,10,10,0.5,0.5,0.5)\) &              &                       \\
\cdashline{2-7}

                                            & \multirow{4}{*}{\((0,0,20,0,0,2)\)} & \multirow{4}{*}{1.90} & \multirow{4}{*}{1.72} & \((0,0,0,0,0,0)\) & \multirow{4}{*}{0.15} & \multirow{4}{*}{0.13} \\
                                            &                                     &                       &                       & \((0,0,5,0,0,0.5)\) &                     &                       \\
                                            &                                     &                       &                       & \((0,0,10,0.5,0.5,1.5)\) &                &                       \\
                                            &                                     &                       &                       & \((10,10,0,1,1,1)\) &                     &                       \\
\cdashline{2-7}

                                            & \multirow{4}{*}{\((0,0,25,0,0,3)\)} & \multirow{4}{*}{2.55} & \multirow{4}{*}{2.30} & \((0,0,0,0,0,0)\) & \multirow{4}{*}{0.21} & \multirow{4}{*}{0.17} \\
                                            &                                     &                       &                       & \((0,0,10,0.5,0.5,1)\) &                  &                       \\
                                            &                                     &                       &                       & \((0,0,15,1,1,1)\) &                      &                       \\
                                            &                                     &                       &                       & \((10,10,0,1.5,1.5,2.5)\) &               &                       \\
\hline
  \end{tabular}
}
\noindent
\parbox{\textwidth}{\footnotesize
\textit{Note:} \((R_x,R_y,R_z,t_x,t_y,t_z)\) denotes the corresponding rotation and translation components, with units of degrees and meters, respectively.
}
\label{table coarse registration impact}
\end{table}

\subsection{Generalizability and Prospective Practical Applications of L2M-Reg}
\label{subsec5.6}

The generalizability of L2M-Reg can be discussed from two perspectives. The first perspective concerns the source of the model data, specifically whether LoD2 models from different agencies can be used. Although L2M-Reg is plane-based and relies on plane information derived from LoD2 models, it is not restricted to models produced by specific agencies or regions. Only the vertex information of each wall surface is required to derive the reference plane parameters. As long as the LoD2 model follows established standards, such as CityGML, and provides access to this information, subsequent processing is independent of the model’s origin.

A fundamental assumption underlying the use of LoD2 models is that building walls are constructed from footprint points that geometrically correspond to the wall plinth. Under this assumption, accurate extraction of the wall plinth area from LiDAR point clouds is essential. This assumption is consistent with common modeling practices across many countries and regions and is therefore considered well-founded. To illustrate this point, the Ingolstadt Store dataset was included as a testbed outside Munich. The LoD2 models and point clouds for this dataset were collected and generated by different agencies than those used for the four Munich datasets, as shown in Table \ref{tabDatasets}. Although the data originate from a different city, all datasets satisfy the footprint point assumption. From a data source perspective, this demonstrates that L2M-Reg is not tied to a specific agency or region and exhibits good generalizability.

The second perspective concerns architectural style, namely whether the method remains applicable to buildings with substantially different designs. This question is closely related to the implementation of the method. L2M-Reg is plane-based, and the algorithm processes facades independently, wall by wall. As a result, its applicability depends on the geometric characteristics of each individual facade. In essence, the key condition is whether a wall surface remains predominantly planar. If this condition is satisfied, this plane-based method remains applicable. If not, the underlying assumption of planarity is violated, and L2M-Reg becomes inapplicable. This limitation is also inherent to many plane-based solutions and represents a clear direction for future improvement.

Regarding the prospective practical applications enabled by L2M-Reg, three key aspects can be summarized:

\begin{enumerate}[label=(\arabic*)]
\item It supports refined, low-cost building-level model updating. It is effective at both the building and the component level. As a result, L2M-Reg substantially reduces data acquisition requirements. High-precision LiDAR-to-Model registration can be achieved by collecting point clouds only in spatially critical areas.

\item It significantly broadens the applicability of LoD2 models in high-precision scenarios. Traditionally, LoD2 models have been used for their large-scale coverage in city-level analyses. Now, L2M-Reg enables accurate LiDAR-to-Model registration. This allows LoD2 models to be reliably used in small-scale, accuracy-sensitive tasks, such as change detection and navigation.

\item It advances the understanding of inherent uncertainty in geospatial data, an aspect often overlooked and considered negligible in LoD2 models. L2M-Reg explicitly connects in-model uncertainty to downstream applications, opening new avenues for uncertainty-aware research and practical applications of geospatial data.
\end{enumerate}

\subsection{Limitations}
\label{subsec5.7}

Although this study employed exclusively real-world datasets for evaluation, given the complexity of real-world scenarios and other unresolved considerations, the current limitations are as follows:

\begin{enumerate}[label=(\arabic*)]
\item As with most fine registration methods, L2M-Reg requires an initial alignment. Developing a one-step solution that removes the dependency on coarse registration will be an important direction for future research. L2M-Reg also assumes that the wall plinth areas are at least partially visible in the point clouds and not completely obstructed by vegetation or vehicles during data acquisition. Although a complete occlusion is very uncommon in practical scenarios, this assumption should be explicitly acknowledged.

\item When an entire facade is composed of transparent glass, most laser returns penetrate the surface and are recorded from interior structures, making it impossible to obtain reliable point clouds of the facade itself. In such cases, L2M-Reg is not applicable. This limitation is inherent to LiDAR-based point cloud acquisition.

\item As a plane-based method, L2M-Reg requires sufficient visibility of planar features in the input point clouds and becomes ineffective when such features are extremely sparse or nearly absent. If a facade is dominated by complex non-planar structures, such as geometrically irregular shapes, fully curved surfaces, arched structures, or wave-like forms, no dominant plane can be reliably identified. In these situations, the underlying assumption of the plane-based logic no longer holds.

\item A complete error propagation model was carefully considered. However, after detailed evaluation, it was determined that a rigorous error propagation analysis cannot be reliably provided at this stage for three main reasons. First, error sources in MLS point clouds are highly complex to model \citep{xu2025pl4u,xu2025propagation}. Second, the complete error modeling of TLS point clouds remains challenging \citep{holst2016}. Third, the accuracy information provided by manufacturers may not be directly applicable. Given these unresolved issues, it is currently not feasible to construct a complete and reliable variance-covariance matrix for the point clouds used in this study. The investigation of the complete variance-covariance matrix is beyond the scope of this study. Relevant details can be found in these publications \citep{wujanz2017,kerekes2020,abdelgafar2025}.
\end{enumerate}

\section{Conclusion}
\label{conclusion}

This paper introduces L2M-Reg, a plane-based LiDAR-to-Model registration method tailored for individual buildings, with a particular focus on addressing the inherent uncertainty in LoD2 models. The key innovations of L2M-Reg are threefold. First, it considers the uncertainties of LoD2 models that serve as reference data, and introduces automated algorithms to identify representative regions and extract plane segments. Second, it introduces the concept of pseudo-planes in GHM and employs a 2D-3D decoupled parameter estimation strategy, which effectively mitigates the influence of low-reliability vertical model data on the estimation of horizontal parameters. Third, it maximizes the use of embedded semantic information in LoD2 models to enhance both the efficiency and reliability of plane correspondence establishment.

L2M-Reg was evaluated against several leading solutions, including both ICP- and plane-based methods, across five real-world datasets. Overall, experimental results show that L2M-Reg outperforms them in both accuracy and computational efficiency. L2M-Reg contributes toward bridging the gap in uncertainty-aware LiDAR-to-Model fine registration at the building level. Moreover, this study enables accurate and cost-efficient model updating at the building level, extends the use of LoD2 models to accuracy-sensitive applications, and explicitly incorporates in-model uncertainty into downstream tasks. Together, this study advances both the methodological understanding and practical exploitation of uncertainty in geospatial data.

Accurate and efficient LiDAR-to-Model registration remains an open research challenge. Further application of L2M-Reg to a wider range of cities and larger-scale datasets, together with tailored optimizations, is planned. In addition, extending L2M-Reg to support seamless registration across both indoor and outdoor environments represents another important direction for future research.

%\end{linenumbers}

\section*{CRediT authorship contribution statement}
\textbf{Ziyang Xu}: Conceptualization, Data curation, Methodology, Software, Resources, Validation, Investigation, Writing – review \& editing, Writing – original draft, Visualization. \textbf{Benedikt Schwab}: Conceptualization, Resources, Data curation, Writing - original draft, Writing – review \& editing. \textbf{Yihui Yang}: Conceptualization, Validation, Visualization, Writing - original draft, Writing – review \& editing. \textbf{Thomas H. Kolbe}: Supervision, Writing - Review \& Editing. \textbf{Christoph Holst}: Supervision, Writing - Review \& Editing, Funding acquisition, Project administration.

\section*{Declaration of competing interest}
The authors declare that they have no known competing financial interests or personal relationships that could have appeared to influence the work reported in this paper.

\section*{Acknowledgments}
This research was funded by TUM Georg Nemetschek Institute of Artificial Intelligence for the Built World, project "NERF2BIM”, PI Christoph Holst. Many thanks to the City of Munich / GeodatenService for supporting this research, to 3D Mapping Solutions\footnote{\url{https://www.3d-mapping.de/}} for providing the MLS point clouds of Ingolstadt as part of the SAVeNoW research project, and to Leonard Trill for data collection.

\section*{Data and code availability}

The datasets used in this paper and the implementation of code for L2M-Reg can be found: \url{https://github.com/Ziyang-Geodesy/L2M-Reg}.

%%\appendix
%%\section{Example Appendix Section}
%%\label{app1}

%%Appendix text.

%%        \citep{<label>} ==> [1]

%%
%%Example citation, See  \citep{lamport94}.

\bibliographystyle{elsarticle-harv} 
\bibliography{main}




% %% For numbered reference style

%   \textit{\LaTeX: a document preparation system},
%   Addison Wesley, Massachusetts,

% \end{thebibliography}

\end{document}